\documentclass{article}
\usepackage{amsfonts}
\usepackage{algorithm}
\usepackage{algorithmic}
\usepackage[ansinew]{inputenc}
\usepackage[pdftex]{graphicx}
\usepackage[cmex10]{amsmath}
\usepackage{url}

\newtheorem{pbm}{Problem}
\newtheorem{thm}{Theorem}

\newtheorem{lem}{Lemma}

\begin{document}

\title{Kernel machines with two layers and multiple kernel learning}

\date{}
\author{Francesco~Dinuzzo\thanks{Francesco~Dinuzzo is with Department of Mathematics, University of Pavia, Pavia, Italy e-mail: francesco.dinuzzo@unipv.it.}}
       
\maketitle

\begin{abstract}%   <- trailing '%' for backward compatibility of .sty file
In this paper, the framework of kernel machines with two layers is introduced, generalizing classical kernel methods. The new learning methodology provide a formal connection between computational architectures with multiple layers and the theme of kernel learning in standard regularization methods. First, a representer theorem for two-layer networks is presented, showing that finite linear combinations of kernels on each layer are optimal architectures whenever the corresponding functions solve suitable variational problems in reproducing kernel Hilbert spaces (RKHS). The input-output map expressed by these architectures turns out to be equivalent to a suitable single-layer kernel machines in which the kernel function is also learned from the data. Recently, the so-called multiple kernel learning methods have attracted considerable attention in the machine learning literature. In this paper, multiple kernel learning methods are shown to be specific cases of kernel machines with two layers in which the second layer is linear. Finally, a simple and effective multiple kernel learning method called RLS2 (regularized least squares with two layers) is introduced, and his performances on several learning problems are extensively analyzed. An open source MATLAB toolbox to train and validate RLS2 models with a Graphic User Interface is available.
\end{abstract}

\section{Introduction} \label{sec01}

Learning by minimizing costs in functional spaces has proven to be an important approach to better understand many estimation problems. Indeed, the functional analytic point of view is the theoretical core of many successful learning methodologies such as smoothing splines, Gaussian processes, and support vector machines \cite{Wahba90, Girosi95, Scholkopf01b, Vapnik98, Shawe-Taylor04, Rasmussen06}, collectively referred to as kernel methods. One of the most appealing properties of kernel methods is optimality according to a variety of representer theorems. These results are usually presented within the theory of RKHS \cite{Aronszajn50}, and formalize the intuition that optimal learning machines trained with a finite number of data must be expressed by a finite number of parameters, even when the hypothesis space is infinite dimensional. Representer theorems have been generalized and analyzed in many forms \cite{Cox90, Scholkopf01, Steinwart03, DeVito04, Micchelli05, Vishwanathan06, Dinuzzo07, Argyriou08}, since their first appearance \cite{Kimeldorf71}.

Recently, has been pointed out that standard kernel machines are somehow limited in their ability to approximate complex functional classes, as a consequence of being shallow architectures. In addition, existing representer theorems only apply to single-layer architectures, though an extension of the theory to include multi-layer networks would be useful to better understand the behavior of current multi-layer architectures and characterize new methods with more flexible approximation capabilities. Such extension is also suggested by complexity theory of circuits \cite{Bengio07} as well as by biological motivated learning models, \cite{Serre07}. In the field of kernel methods, the need for complex hypothesis spaces reflecting “broad” prior knowledge has led to the idea of learning the kernel from empirical data simultaneously with the predictor \cite{Bach04,Lanckriet04,Ong05,Micchelli05b,Argyriou05,Wu07,Micchelli07}. Indeed, the difficulty of choosing a good hypothesis space with little available a-priori knowledge is significantly reduced when the kernel is also learned from the data. The flexibility of algorithms implementing the framework of kernel learning makes also possible to address important machine learning issues such as feature selection, learning from heterogeneous sources of data, and multi-scale approximation. A major extension to the framework of classical kernel methods, based on the concept of \emph{hyper-kernels}, has been introduced in \cite{Ong05}, encompassing many convex kernel learning algorithms. In this paper, the connection between \emph{learning the kernel} in standard (single layer) kernel machines and learning in \emph{multi-layer architectures} is analyzed. We introduce the framework of \emph{kernel machines with two layers}, that also encompasses classical single layer kernel methods as well as many kernel learning algorithms.

Consider a generic architecture whose input-output behavior can be described as a function composition of two \emph{layers}
\begin{equation}\label{E01}
f = f_2 \circ f_1, \quad f_1: X \rightarrow Z, \quad f_2: Z \rightarrow Y,
\end{equation}
\noindent where $X$ is a generic set while $Z$ and $Y$ are two Hilbert spaces. The problem of learning simultaneously the two layers $f_1$ and $f_2$ from input-output data pairs $(x_i,y_i) \in X \times Y$ can be formalized in a functional analytic setting. Indeed, in section \ref{sec02} it is shown that a representer theorem holds: even if $f_1$ and $f_2$ are searched into infinite dimensional function spaces, optimal learning architectures are finite linear combination of kernel functions on each layer, where optimality is measured according to general regularization functionals. Remarkably, such representer theorem also imply that, upon training, architecture (\ref{E01}) can be equivalently regarded as a standard kernel machine in which the kernel function has been learned from the data. After discussing the general result on the solution representation for two non-linear layers, the attention is focused on the case in which the second layer is linear. In section \ref{sec03}, we introduce a regularization framework that turns out to be equivalent to a general class of methods to perform \emph{multiple kernel learning}, that is the simultaneous supervised learning of a predictor and the associated kernel as a convex combination of \emph{basis kernels}. The general problem of \emph{learning the kernel} is receiving a lot of attention in recent years, both from functional analytic point of view and from pure optimization perspectives. Since the earlier works \cite{Lanckriet04, Micchelli05b}, many improved optimization schemes have been proposed \cite{Argyriou06,Sonnenburg06, Micchelli07, Rakotomamonjy08, Bach08}. In section \ref{sec04}, a method called RLS2 (regularized least squares with two layers) based on regularized multiple kernel learning with the square loss function is introduces and studied. Along the line of recent advances in multiple kernel learning \cite{Sonnenburg06, Rakotomamonjy08, Bach08}, it is shown that the involved optimization can be efficiently carried out using a two-step procedure. For RLS2, two-step optimization turns out to be especially simple and computationally appealing, alternating between the solution of a linear system and a constrained least squares problem. The application of RLS2 on a variety of learning problems is analyzed in section \ref{sec05}. State of the art generalization performances are achieved on several datasets, including multi-class classification of genomic data. An open source MATLAB toolbox to train and validate RLS2 models with a Graphic User Interface is available at \url{ http://www.mloss.org}. All the proof of Theorems and Lemmas are given in the Appendix.

\section{A representer theorem for architectures with two layers} \label{sec02}

A learning architecture with two layers can be formalized as a map $f: X \rightarrow Y$ expressed as a function composition as in equation (\ref{E01}). Introduce two RKHS $\mathcal{H}_1$ and $\mathcal{H}_2$ of vector-valued functions \cite{Micchelli05} defined over $X$ and $Z$ respectively, with (operator-valued) kernel functions $K^1$ and $K^2$, and consider the following problem:

\begin{pbm}\label{PBM01}
\[
\min_{\substack{f_1 \in \mathcal{H}_1,\\ f_2 \in\mathcal{H}_2} } \sum_{i=1}^{\ell}L_i\left((f_2 \circ f_1)(x_i)\right) +R_1(\|f_1 \|_{\mathcal{H}_1})+R_2(\|f_2 \|_{\mathcal{H}_2}).
\]
\end{pbm}

Here, $L_i:Y\rightarrow\mathbb{R}_+$ are loss functions measuring the approximation of training data, while $R_1, R_2:\mathbb{R}_+ \rightarrow \mathbb{R}_+ \cup \{+\infty\}$ are two extended-valued non-decreasing functions (not identically $+\infty$) that play the role of regularization terms. Problem \ref{PBM01} is outside the scope of standard representer theorems \cite{Scholkopf01} due to the presence of the composition $(f_2 \circ f_1)$. Nevertheless, it still holds that linear combinations of a finite number of kernel functions are optimal solutions, as soon as there exist minimizers.

\begin{thm}\label{THM1}
If the functional of Problem \ref{PBM01} admit minimizers, then there exist optimal solutions of Problem \ref{PBM01} in the form
\[
f_1(x) =  \sum_{i=1}^{\ell} K^1(x,x_i)a_{i},\qquad f_2(z) = \sum_{i=1}^{\ell} K^2(f_1(x_i),z) b_{i}.
\]
\noindent Therefore, there exists optimal learning architectures in the following input-output form:
\begin{equation}\label{E02}
f(x) = (f_2 \circ f_1)(x) = \sum_{i=1}^{\ell} K(x_i,x)b_{i}, \qquad K(x_1,x_2) := K^2(f_1(x_1), f_1(x_2)).
\end{equation}
\end{thm}

\noindent Theorem \ref{THM1} is a \emph{restriction} theorem: the search for solutions of Problem \ref{PBM01} can be restricted to \emph{kernel machines with two layers} involving a finite number of kernel functions, even when $\mathcal{H}_1$ and $\mathcal{H}_2$ are infinite dimensional spaces. Notice that Theorem \ref{THM1} is not an existence theorem, since existence of minimizers is one of the hypotheses. As shown in the next sections, existence can be ensured under mild additional conditions on $L_i, R_1, R_2$. Under the general hypothesis of Theorem \ref{THM1}, uniqueness of minimizers in Problem \ref{PBM01} is also not guaranteed, even when loss functions $L_i$ are strictly convex. Notice also that Theorem \ref{THM1} do admit the presence of optimal solutions not in the form of finite kernel expansions. However, if such solutions exist, then their projections over the finite dimensional span of kernel sections are optimal as well, so that one can restrict the attention to kernel machines with two layers also in this case. Finally, when $R_1$ and $R_2$ are strictly increasing, it holds that \emph{every} optimal solution of Problem \ref{PBM01} can be expressed as a kernel machine with two layers.

\section{Multiple kernel learning as a kernel machine with two layers} \label{sec03}

Theorem \ref{THM1} shows that training an architecture with two layers is equivalent to train simultaneously a single-layer kernel network and the kernel function, see equation (\ref{E02}). In this section, it is shown that \emph{multiple kernel learning}, consisting in simultaneous learning of a finite linear combination of kernels and the associated predictor, can be interpreted as a specific instance of kernel architecture with two layers. Introduce a set of $m$ positive kernels $\widetilde{K}_{i}$ defined on $X \times X$, called  \emph{basis kernels} and consider the following choice for $\mathcal{H}_1$ and $\mathcal{H}_2$.

\begin{itemize}
   \item $\mathcal{H}_1$ is an RKHS of vector valued functions $f: X \rightarrow \mathbb{R}^m$ associated with the matrix-valued kernel function $K^1$ such that
    \[
    K^1(x_1,x_2) = \textrm{diag}\left\{\widetilde{K}_{1}(x_1,x_2), \ldots, \widetilde{K}_{m}(x_1,x_2) \right\}.
    \]

   \item  $\mathcal{H}_2$ is the RKHS of real valued functions $f: \mathbb{R}^m \rightarrow \mathbb{R}$ associated with the linear kernel
   \[
   K_2\left(z_1,z_2\right) = z_1^TSz_2,
   \]
   \noindent where $S$ is a diagonal scaling matrix:
   \[
   S = \textrm{diag}\left\{s_1, \ldots, s_{m}\right\} > 0.
   \]
\end{itemize}

\noindent For any $f \in \mathcal{H}_1$, let $f^i$, ($i = 1, \ldots, m$) denote its components. Introduce the indicator function $I$ of the interval $[0, 1]$ defined as
\[
I(x) = \left\{
\begin{array}{ll}
0, & 0 \leq x \leq 1 \\
+\infty, & x > 1
\end{array}
\right.,
\]

\noindent let $L_i:Y\rightarrow \mathbb{R}_+$ denote lower semi-continuous convex loss functions and $\lambda > 0$. In the following, we analyze the particular case of Problem \ref{PBM01} in which $R_1$ is a square regularization and $R_2$ is the indicator function regularization:
\[
R_1(x) = \frac{\lambda}{2} x^2, \qquad R_2(x) = I(x).
\]

\begin{pbm}\label{PBM02}
\[
\min_{\substack{f_1 \in \mathcal{H}_1,\\ f_2 \in\mathcal{H}_2} } \left[\sum_{i=1}^{\ell}L_i\left((f_2\circ f_1)(x_i)\right) + \frac{\lambda}{2} \| f_1 \|^2_{\mathcal{H}_1} + I\left(\| f_2 \|_{\mathcal{H}_2} \right) \right].
\]
\end{pbm}

\noindent Let's briefly discuss the choice of regularizers. First of all, notice that both $R_1$ and $R_2$ are convex functions. Since $L_i$ are convex loss functions and $f_2$ is linear, the problem is separately convex in both $f_1$ and $f_2$. Apparently, regularizing with the indicator function $I\left(\| f_2 \|_{\mathcal{H}_2}\right)$ is equivalent to impose the constraint $\| f_2 \|_{\mathcal{H}_2} \leq 1$. Lemma \ref{LEM1} below shows that minimization into the unitary ball can be carried out without any loss of generality.

\begin{lem}\label{LEM1}
Let $(f_1^*, f_2^*)$ denote an optimal solution of the following problem:
\[
\min_{\substack{f_1 \in \mathcal{H}_1,\\ f_2 \in\mathcal{H}_2} } \left[\sum_{i=1}^{\ell}L_i\left((f_2\circ f_1)(x_i)\right) + \frac{\alpha}{2} \| f_1 \|^2_{\mathcal{H}_1} + \gamma \cdot I\left(\frac{\| f_2 \|_{\mathcal{H}_2}}{\beta} \right) \right].
\]

\noindent Then, $(f_1, f_2) = (\beta f_1^*, f_2^* / \beta)$ is an an optimal solution of Problem \ref{PBM02} with $\lambda = \alpha/\beta^2$ and satisfy
\begin{equation}\label{E15}
f = f_2 \circ f_1 = f_2^* \circ f_1^*.
\end{equation}

\end{lem}

Thanks to the scaling properties coming from the linearity of the second layer and the use of the indicator function, the introduction of an additional regularization parameter can be avoided, thus significantly reducing the complexity of model selection. The next Theorem characterizes optimal solutions of Problem \ref{PBM02}.

\begin{thm}\label{THM2}
There exist optimal solutions $f_1$ and $f_2$ of Problem \ref{PBM02} in the form
\[
f_1^i(x) = s_iw_i\sum_{j=1}^{\ell}c_j\widetilde{K}_i(x_j,x), \qquad f_2(z) = z^T S w.
\]
\noindent Letting $d_i := s_iw_i^2$, optimal coefficients $(c,d)$ solves the multiple kernel learning Problem \ref{PBM07} below, where
\begin{equation}\label{E10}
Q(z) := \sum_{j=1}^{\ell}L_j(z_j).
\end{equation}
\noindent Finally, the solution of Problem \ref{PBM02} can be written as in equation (\ref{E02}), where the kernel $K$ satisfies
\begin{equation}\label{E03}
K(x,y) = \sum_{i=1}^{m} d_i K_i(x,y), \qquad K_i(x,y) =  \sum_{j_1=1}^{\ell}\sum_{j_2=1}^{\ell}c_{j_1}c_{j_2} \widetilde{K}_i(x_{j_1},x)\widetilde{K}_i(x_{j_2},y).
\end{equation}
\end{thm}

\begin{pbm}\label{PBM07}
\[
\min_{c \in \mathbb{R}^{\ell},d \in \mathbb{R}^{m}}\left(Q(R(d)c) + \frac{\lambda}{2}c^T R(d) c \right)
\]
\noindent subject to
\begin{equation}\label{E06}
R^k_{ij} = s_k \widetilde{K}_k(x_i,x_j), \qquad R(d) = \sum_{k=1}^{m} d_k R^k, \qquad d_k \geq 0, \qquad \sum_{k=1}^{m}d_k \leq 1.
\end{equation}

\end{pbm}

Theorem \ref{THM2} shows that the variational Problem \ref{PBM02} for a two-layer kernel machine is equivalent to the multiple kernel learning Problem \ref{PBM07}. The non-negativity constraints $d_k \geq 0$ leads to a sparse selection of a subset of basis kernels. In standard formulations, multiple kernel learning problems feature the equality constraint $\sum_{k=1}^{m}d_k =1$, instead of the inequality in (\ref{E06}). Nevertheless, Lemma \ref{LEM2} below shows that there always exist optimal solutions of Problem \ref{PBM03} satisfying the equality, so that the two optimization problems are equivalent.

A few comments on certain degeneracies in Problem \ref{PBM02} are in order. First of all, observe that the absolute value of optimal coefficients $w_i$ characterizing the two layer kernel machine is given by $|w_i| = \sqrt{d_i/s_i}$, but $\textrm{sign}(w_i)$ is undetermined. Then, without loss of generality, it is possible to choose $w_i = \sqrt{d_i/s_i}$. Second, observe that the objective functional of Problem \ref{PBM07} depends on $c$ through the product $R c$. When $R$ is singular, the optimal vector $c$ is not unique (independently of $Q$). In particular, if $v$ belongs to the null space of $R$, then $c+\gamma v$ achieves the same objective value of $c$, for any $\gamma \in \mathbb{R}$. This case also occurs in standard (single-layer) kernel methods. One possible way to break the indetermination, again without any loss of generality, is to constrain $c$ to belong to the range of $R$. With such additional constraint, there exists $z$ such that
\begin{equation}\label{E20}
c = R^{\dag} z,
\end{equation}
where $\dag$ denote the Moore-Penrose pseudo-inverse (notice that, in general, $z$ might be different from $Rc$). Remarkably, the introduction of such change of variable makes also possible to derive an addition formulation of Problem \ref{PBM07}, which can be shown to be a convex optimization problem. Indeed, by rewriting Problem \ref{PBM07} as a function of $(z,d)$, the following problem is obtained:

\begin{pbm}\label{PBM03}
\[
\min_{z \in \mathbb{R}^{\ell},d \in \mathbb{R}^{m}}\left(Q(z) + \frac{\lambda}{2}z^T R^{\dag}(d) z \right), \quad \textrm{ subject to } \quad (\ref{E06}).
\]
\end{pbm}

\begin{lem}\label{LEM2}
Problem \ref{PBM03} is a convex optimization problem and there exists an optimal vector $d$ satisfying the equality constraint
\begin{equation}\label{E07}
\sum_{k=1}^{m}d_k = 1.
\end{equation}
\end{lem}

Lemma \ref{LEM2} completes the equivalence between the specific \emph{kernel machines with two layers} obtained by solving Problem \ref{PBM02} and \emph{multiple kernel learning} algorithms. The Lemma also gives another important insight into the structure of Problem \ref{PBM07}: local minimizers are also global minimizers, a property that directly transfer from Problem \ref{PBM03} through the change of variable (\ref{E20}).

\subsection{Linear machines} \label{sec03.1}

In applications of standard kernel methods involving high-dimensional input data, the linear kernel on $\mathbb{R}^N$
\begin{equation}\label{E13}
K(x_1,x_2) = x_1^T x_2
\end{equation}
\noindent plays an important role. Optimization algorithms for linear machines are being the subject of a renewed attention in the literature, due to some important experimental findings. First, it turns out that linear models are already enough flexible to achieve state of the art classification performances in application domains such as text document classification, word-sense disambiguation, and drug design, see e.g. \cite{Joachims06}. Second, linear machines can be trained using extremely efficient and scalable algorithms \cite{Hsieh06,Shalev-Shwartz07,Fan08}. Finally, linear methods can be also used to solve certain non-linear problems (by using non-linear feature maps), thus ensuring a good trade-off between flexibility and computational convenience.

Linear kernels are also meaningful in the context of multiple kernel learning methods. Indeed, when the input set $X$ is a subset of $\mathbb{R}^N$, a possible choice for the set of basis kernels $\widetilde{K}_k$ is given by linear kernels on each component:
\begin{equation}\label{E12}
\widetilde{K}_k(x_1,x_2) = x_1^kx_2^k.
\end{equation}
\noindent Such a choice makes the input output map (\ref{E02}) a linear function:
\begin{equation}\label{E16}
f(x) = \sum_{j=1}^{m} \left(d_j s_j \sum_{i=1}^{\ell}c_ix_i^j\right) x^j = a^T x,
\end{equation}
\noindent where
\begin{equation}\label{E17}
a_j := d_j s_j  z_j, \qquad z_j := \sum_{i=1}^{\ell}c_ix_i^j.
\end{equation}

\noindent Here, an important benefit is sparsity in the vector of weights $a$, that follows immediately from sparsity of vector $d$. In this way, linear multiple kernel learning algorithms can simultaneously perform regularization and linear \emph{feature selection}. Such property is apparently linked to the introduction of the additional layer in the architecture, since standard kernel machines with one layer are not able to perform any kind of automatic feature selection. From the user's point of view, linear kernel machines with two layers behave similarly to sparse $\ell_1$ regularization methods such as the Lasso \cite{Tibshirani96}, performing feature selection by varying with continuity a shrinking parameter. However, it seems that $\ell_1$ regularization methods cannot be interpreted as kernel machines (not even with two layers) and these two classes of algorithms are thus distinct. An instance of linear regularization methods with two layers is proposed in subsection \ref{sec04.2} and analyzed in the experimental section \ref{sec05}.

\section{Regularized least squares with two layers}  \label{sec04}

\begin{algorithm}
\caption{Alternate optimization for RLS2} \label{ALG1}
\begin{algorithmic}
\STATE $i \leftarrow \arg\max_{k=1,\ldots,m} y^T R^k y$
\STATE $d \leftarrow e_i$
\STATE $B \leftarrow \{i\}$
\WHILE{(stopping criterion is not met)}
\STATE $R \leftarrow 0$
\FOR{$j  \in B$}
\STATE $R \leftarrow R + d_j R^j$
\ENDFOR
\STATE $c \leftarrow \textrm{Solution of the linear system }\left(R+\lambda I\right)c = y$
\STATE $u \leftarrow \left(y-\frac{\lambda c}{2}\right)$
\FOR{$i =1, \ldots, m$}
\STATE $v_i \leftarrow R^i c$
\ENDFOR
\STATE $d \leftarrow$ Solution of Problem (\ref{PBM06}).
\STATE $B \leftarrow \left\{j: d_j \neq 0 \right\}$
\ENDWHILE
\end{algorithmic}
\end{algorithm}

In the previous section, a general class of convex optimization problems to learn finite linear combinations of kernels is shown to be equivalent to a two-layer kernel machine. As for standard kernel machines, different choices of loss functions $L_i$ lead to a variety of learning algorithms. For instance, from the results of the previous section it follows that the two-layer version of standard Support Vector Machines with “hinge” loss functions $L_i(z) = \left(1-y_iz\right)_{+}$ is equivalent to the SILP (Semi-Infinite Linear Programming) multiple kernel learning problem studied in \cite{Sonnenburg06}, whose solution can be computed, for instance, by using gradient descent or SimpleMKL \cite{Rakotomamonjy08}.

In this section, attention is focussed on square loss functions $L_i(z) = (y_i-z)^2/2$ and the associated kernel machine with two layers. As we shall show, coefficients $c_j$ and $d_j$ defining the architecture as well as the “equivalent input-output kernel” $K$ can be computed by solving a very simple optimization problem. Such problem features the minimization of a \emph{quartic} functional in $(c,d)$, that is separately quadratic in both $c$ and $d$. It is worth noticing that the square loss function can be used to solve regression problems as well as classification ones. Indeed, generalization performances of regularized least squares classifiers have been shown to be comparable to that of Support Vector Machines on many dataset, see \cite{Rifkin03, Fung05} and references therein.

\begin{pbm} [Regularized least squares with two layers (RLS2)] \label{PBM04}
\[
\min_{c \in \mathbb{R}^{\ell},d \in \mathbb{R}^{m}}\left(\frac{1}{2}\left\|y- R(d)c\right\|^2 + \frac{\lambda}{2}c^T R(d) c \right), \quad \textrm{ subject to } \quad (\ref{E06}).
\]
\end{pbm}

\noindent Let $\Delta_m$ denote the standard $(m-1)$-simplex in $\mathbb{R}^m$:
\[
\Delta_m:=\left\{ d \in \mathbb{R}^m: \quad d \geq 0,\quad \sum_{i=1}^{m}d_i = 1\right \}.
\]
\noindent For any fixed $d$, Problem \ref{PBM04} is an unconstrained quadratic optimization problem with respect to $c$. It is then possible to solve for the optimal $c^*$ in closed form as a function of $d$:
\begin{equation}\label{E09}
c^*(d) = \left(\sum_{i=1}^{m}d_iR^i  + \lambda I\right)^{-1} y.
\end{equation}

\noindent As shown in Lemma \ref{LEM3} below, Problem \ref{PBM04} can be reduced to the following Problem in $d$ only.

\begin{pbm}\label{PBM05}
\[
\min_{d  \in \Delta_m}\frac{\lambda}{2}y^{T} c^*(d),
\]
\end{pbm}

\begin{lem}\label{LEM3}
The pair $(c^*,d^*)$ is an optimal solution of Problem \ref{PBM04} if and only if equation (\ref{E09}) holds and $d^*$ is an optimal solution of Problem \ref{PBM05}.
\end{lem}

Along the lines of recent developments in multiple kernel learning optimization \cite{Sonnenburg06, Rakotomamonjy08}, we propose a two-step minimization procedure that alternates between kernel and predictor optimization. The specific structure of our problem allows for exact optimization in each of the two phases of the optimization process. Let
\[
V  := \left(
\begin{array}{ccc}
v_1 & \cdots & v_m \\
\end{array}
\right) = \left(
\begin{array}{ccc}
R^1 c & \cdots & R^m c \\
\end{array}
\right),\qquad u  :=  \left(y-\frac{\lambda c}{2}\right).
\]

\noindent For any fixed $c$, minimization with respect to $d$ boils down to the following simplex-constrained least squares problem, as ensured by the subsequent Lemma \ref{LEM4}.

\begin{pbm}\label{PBM06}
\[
\min_{d  \in \Delta_m}\|V d-u\|^2.
\]
\end{pbm}

\begin{lem}\label{LEM4}
For any fixed $c$, the optimal coefficient vector $d$ of Problem \ref{PBM04} can be obtained as the solution of Problem \ref{PBM06}.
\end{lem}

\noindent Optimal coefficients can be computed using an iterative two-step procedure such as the Algorithm \ref{ALG1}, that alternates between minimization with respect to $c$ obtained through the solution of the linear system (\ref{E09}), and the solution of the simplex-constrained least squares Problem  \ref{PBM06} in $d$. The non-negativity constraint induce sparsity in the vector $d$, and thus also in the input-output kernel expansion. To understand the initialization of coefficients $c$ and $d$ in Algorithm \ref{ALG1}, consider the limiting solution of the optimization problem when the regularization parameter tends to infinity. Such solution is the most natural starting point for a regularization path because optimal coefficients can be computed in closed form.

\begin{lem}\label{LEM5}
The limiting solution of Problem \ref{PBM04} when $\lambda \rightarrow +\infty$ is given by
\[
(c_{\infty}, d_{\infty}) = \left(0, e_i\right),\qquad i \in \arg\max_{k=1,\ldots,m} y^T R^k y.
\]
\end{lem}

\noindent As shown in subsection \ref{sec04.3}, the result of Lemma \ref{LEM5} can be also used to give important insights into the choice of the scaling $S$ in the second layer.

\subsection{A Bayesian maximum a posteriori interpretation of RLS2}  \label{sec04.1}

The equivalence between regularization Problem \ref{PBM02} and the multiple kernel learning optimization Problem \ref{PBM08} can be readily exploited to give a Bayesian MAP (maximum a posteriori) interpretation of RLS2. To specify the probabilistic model, we need to put a prior distribution over the set of functions from $X$ into $\mathbb{R}$ and define the data generation model (likelihood). In the following, $N(\mu,\sigma^2)$ denote a real Gaussian distribution with mean $\mu$ and variance $\sigma^2$, $GM(f,K)$ a Gaussian measure on the set of functions from $X$ into $\mathbb{R}^m$ with mean $f$ and covariance function $K$, and $U(\Omega)$ the uniform distribution in $\mathbb{R}^m$ over a set $\Omega$ of positive finite measure. Let $f: X \rightarrow \mathbb{R}$ be such that
\[
f = w^T S f_1,
\]
\noindent where $f_1:X\rightarrow \mathbb{R}^m$ is distributed according to a Gaussian measure over $\mathcal{H}_1$ with zero mean and (matrix-valued) covariance function $K^1$, and $w$ is a random vector independent of $f_1$, distributed according to an uniform distribution over the ellipsoid $E_S := \{w\in\mathbb{R}^m: w^TSw \leq 1\}$:
\[
f_1 \sim GM\left(0,K^1\right), \qquad w \sim U(E_S).
\]

\noindent Regarding the likelihood, assume that the data set $\mathcal{D} := \{(x_i,y_i)\}_{i=1}^{\ell}$, is generated by drawing pairs $(x_i,y_i)$ independently and identically distributed according to the usual additive Gaussian noise model:
\[
y_i|(x_i,f) \sim N(f(x_i),\sigma^2).
\]

\noindent When $\mathcal{H}_1$ is a finite dimensional space, the MAP estimate of $f$ is:
\[
f^{*} = w^{*T} S f_1^{*},
\]
\noindent where $\left(f_1^{*},w^{*}\right)$ maximize the posterior density:
\[
p(f| \mathcal{D}) \propto p(\mathcal{D}|f) p(f_1) p(w).
\]

\noindent Specifically, we have
\begin{eqnarray*}
p(\mathcal{D}|f) & \propto & \prod_{i=1}^{\ell} \textrm{exp}\left(-\frac{\left(y_i-w^T S f_1(x_i)\right)^2}{2\sigma^2}\right),\\
p(f_1) & \propto & \textrm{exp}\left(-\frac{\|f_1\|^2_{\mathcal{H}_1}}{2}\right),\\
p(w) & \propto &\left\{
                    \begin{array}{ll}
                      1, & x \in E_S\\
                      0, & \hbox{else}
                    \end{array}
                  \right..
\end{eqnarray*}

\noindent It follows that $w^{*} \in E_S$ and, by taking the negative logarithm of $p(f|\mathcal{D})$, that the MAP estimate coincides with the solution of the regularization Problem \ref{PBM02} with square loss functions and $\lambda = \sigma^2$. When $\mathcal{H}_1$ is an infinite-dimensional function space, the Gaussian measure prior for $f_1$ do not admit a probability density. Nevertheless, the regularization Problem \ref{PBM02} can be still recovered by understanding MAP estimate as a maximal point of the posterior probability measure, as described in \cite{Hengland07}.

\subsection{Linear regularized least squares with two layers} \label{sec04.2}

As described in subsection \ref{sec03.1}, when the input set $X$ is a subset of $\mathbb{R}^m$ the linear choice of basis kernels (\ref{E12}) produces linear machines with feature selection capabilities. First of all, recall that standard regularized least squares with the linear kernel (\ref{E13}) boils down to finite-dimensional Tikhonov regularization \cite{Tikhonov77} also known as ridge regression \cite{Hoerl70}:
\[
\min_{a \in \mathbb{R}^m} \left( \|y-H a \|^2 + \lambda \|a\|^2 \right),
\]
\noindent where $H \in \mathbb{R}^{\ell \times m}$ denote the matrix of inputs such that $H_{ij} = x_i^j$. Lemma \ref{LEM6} below states that the linear version of RLS2 is equivalent to a “scaled” ridge regression problem, in which the optimal scaling is also estimated from the data. Let
\begin{equation}\label{E18}
\Gamma(d) := \textrm{diag}\left\{s_1 d_1, \ldots, s_m d_m\right\}.
\end{equation}

\noindent For any fixed $d$, let $n(d)$ be the number of non-zero coefficients $d_i$, $\widetilde{\Gamma}(d) \in \mathbb{R}^{n(d) \times n(d)}$ denote the diagonal sub-matrix of $\Gamma$ containing all the strictly positive coefficients. Moreover, let $\widetilde{H}$ denote the scaled sub-matrix of selected features $\widetilde{H} := H \widetilde{\Gamma}$.

\begin{pbm}\label{PBM08}
\[
\min_{\substack{\widetilde{z} \in \mathbb{R}^{n(d)},\\ d  \in \Delta_m}}\|y-\widetilde{H} \widetilde{z} \|^2 + \lambda \|\widetilde{\Gamma}^{1/2}(d)\widetilde{z}\|^2,\quad \textrm{ subject to } \quad (\ref{E18}).
\]
\end{pbm}

\begin{lem}\label{LEM6}
When basis kernels are as in (\ref{E12}), the optimal solution of Problem \ref{PBM04} can be written as in (\ref{E16})-(\ref{E17}), where $(z,d)$ solves Problem \ref{PBM08}.
\end{lem}

\noindent  From Problem \ref{PBM08}, one can easily see that when $d$ is fixed to its optimal value, the optimal $\widetilde{z}$ in Problem \ref{PBM08} is given by the familiar expression:
\[
\widetilde{z} = \left(\widetilde{H}^T\widetilde{H}+\lambda \widetilde{\Gamma}\right)^{-1}\widetilde{H}^T y.
\]
\noindent The result of Lemma \ref{LEM6} can be also used to give an interesting interpretation of linear RLS2. In fact, the coefficient $s_id_i$ can be interpreted as a quantity proportional to the inverse variance of the $i$-th coefficient $z_i$. Hence, Problem \ref{PBM08} can be seen as a Bayesian MAP estimation with Gaussian residuals, Gaussian prior on the coefficients and uniform prior over a suitable simplex on the vector of inverse coefficients's variances.

It is useful to introduce a notion of “degrees of freedom”, see e.g. \cite{Efron04,Hastie08}. Degrees of freedom is an index more interpretable than the regularization parameter, and can be also used to choose the regularization parameter according to tuning criteria such as $C_p$ \cite{Mallows73}, $\textrm{AIC}$ \cite{Akaike73}, $\textrm{BIC}$ \cite{Schwarz78}, $\textrm{GCV}$ \cite{Craven79}. A general expression for the effective degrees of freedom of non-linear kernel regression methods with one layer, based on the SURE (Stein's Unbiased Risk Estimator) approximation \cite{Stein81} has been recently derived in \cite{Dinuzzo08}. For linear RLS2, the following quantity seems an appropriate approximation of the degrees of freedom:
\begin{equation}\label{E14}
\widehat{df}(\lambda) = \textrm{tr}\left(\widetilde{H}\left(\widetilde{H}^T\widetilde{H}+\lambda \widetilde{\Gamma}\right)^{-1}\widetilde{H}^T\right).
\end{equation}
\noindent Expression (\ref{E14}) corresponds to the equivalent degrees of freedom of a linear regularized estimator with regressors fixed to $\widetilde{H}$ and diagonal regularization $\lambda \widetilde{\Gamma}$. Notice that (\ref{E14}) neglects the non-linear dependence of matrix $\widetilde{H}$ on the output data and does not coincide with the SURE estimator of the degrees of freedom. Nevertheless, the property $0 \leq \widehat{df}(\lambda) \leq m$ holds, so that $\widehat{df}$ can be conveniently used to interpret the complexity of the linear RLS2 model (see subsection \ref{sec05.1} for an example).

\subsection{Choice of the scaling and feature/kernel selection} \label{sec04.3}

The ability of RLS2 to select features or basis kernels is highly influenced by the scaling $S$ in the second layer. In this subsection, we analyze certain scaling rules that are connected with popular statistical indices, often used as “filters” for feature selection \cite{Guyon06}. Since the issue of scaling still needs further investigation, it is not excluded that new rules different from those mentioned in this subsection may work better on specific problems.

A key observation is the following: according to Lemma \ref{LEM5}, RLS2 with heavy regularization favors basis kernels that maximizes the quantity $A_k = y^TR^ky$, that represents a kind of \emph{alignment} between the kernel $R^k$ and the outputs. This means that RLS2 tends to select kernels that are highly aligned with the outputs. Since each alignment $A_k$ is proportional to the scaling factor $s_k$, an effective way to choose the scaling is one that makes the alignment a meaningful quantity to maximize. First of all, we discuss the choice of scaling for the linear RLS2 algorithm introduces in subsection \ref{sec04.2}. The generalization to the case of non-linear basis kernels easily follows by analyzing the associated feature maps.

In the linear case, we have $R^k = s_k x^k x^{kT}$, where $x^k$ is the $k$-th feature vector, so that
\[
A_k = s_k (y^Tx^k)^2.
\]
\noindent By choosing
\[
s_k = (\|y\|\|x^k\|)^{-2},
\]
\noindent the alignment becomes the squared cosine of the angle between the $k$-th feature vector and the output vector:
\[
A_k=\left(\frac{y^Tx^k}{\|y\|\|x^k\|}\right)^2 = \cos^2 \theta_k.
\]
\noindent In particular, when the outputs $y$ and the features $x^k$ are centered to have zero mean, $A_k$ coincides with the squared Pearson correlation coefficient between the outputs and the $k$-th feature, also known as \emph{coefficient of determination}. This means that RLS2 with heavy regularization selects the features that mostly correlates to the outputs. Since the term $\|y\|^2$ is common to all the factors, one can also use
\begin{equation}\label{E21}
s_k = \|x^k\|^{-2},
\end{equation}
\noindent without changing the profile of solutions along a regularization path (though, the scale of regularization parameters is shifted). Observe that rule (\ref{E21}) may also make sense when data are not centered or centered around values other than the mean. In fact, for some datasets, performances are better without any centering (this is the case, for instance, of Experiment 1 in subsection \ref{sec05.1}). Notice also that (\ref{E21}) only uses training inputs whereas, in a possible variation, one can replace $x^k$ with the vector containing values of the $k$-th feature for both training and test data (when available). The latter procedure sometimes work better than scaling using training inputs only, and will be referred to as \emph{transductive scaling} in the following. For binary classification with labels $\pm 1$, the choice (\ref{E21}) with or without centering still make sense, but other rules are also possible. Let $\ell_+$ and $\ell_-$ denote the number of samples in the positive and negative class, respectively, and $m^k_+$ and $m^k_-$ denote the within-class mean values of the $k$-th feature:
\[
m^k_+ = \frac{1}{\ell_+}\sum_{i:y_i=1}x_i^k, \qquad m^k_- = \frac{1}{\ell_-}\sum_{i:y_i=-1}x_i^k.
\]
\noindent By choosing
\begin{equation}\label{E22}
s_k = \frac{1}{(\sigma_+^k)^2+(\sigma_-^k)^2},
\end{equation}
\noindent where $\sigma_+^k$ and $\sigma_-^k$ denote the within class standard deviations of the $k$-th feature, one obtain
\[
A_k = \frac{(\ell_+ m^k_+-\ell_-m^k_-)^2}{(\sigma_+^k)^2+(\sigma_-^k)^2}.
\]
\noindent When the two classes are balanced ($\ell_+ = \ell_- = \ell/2$), $A_k$ boils down to a quantity proportional to the classical Fisher criterion (or signal-to-interference ratio):
\[
A_k = \frac{\ell^2}{4}\frac{(m^k_+-m^k_-)^2}{(\sigma_+^k)^2+(\sigma_-^k)^2}.
\]

Rules (\ref{E21}) and (\ref{E22}) can be generalized to the case of non-linear basis kernels, by observing that non-linear kernels can be always seen as linear upon mapping the data in a suitable \emph{feature space}. A rule that generalizes (\ref{E21}) is the following \cite[see e.g.][]{Rakotomamonjy08}:
\begin{equation}\label{E11}
s_k = \left(\sum_{i=1}^{\ell}\widetilde{K}_k(x_i,x_i)\right)^{-1},
\end{equation}
\noindent that amounts to scale each basis kernel by the trace of the kernel matrix, and reduces exactly to (\ref{E21}) in the linear case. Also (\ref{E11}) can be applied with or without centering. A typical centering is \emph{normalization in feature space}, that amounts to subtract $1/\ell\sum_{i,j}\widetilde{K}_k(x_i,x_j)$ to the basis kernel $\widetilde{K}_k$, before computing (\ref{E11}). A transductive scaling rule can be obtained by extending the sum to both training and test inputs, namely computing the inverse trace of the overall kernel matrix, as in \cite{Lanckriet04}. Finally, a non-linear generalization of (\ref{E22}) is given by the following:
\begin{multline*}
s_k = \bigg[\sum_{i:y_i=1}^{\ell}\left(\frac{\widetilde{K}_k(x_i,x_i)}{\ell_+}-\sum_{j:y_j=1}\frac{\widetilde{K}_k(x_i,x_j)}{\ell_+^2}\right) +\\ \sum_{i:y_i=-1}^{\ell}\left(\frac{\widetilde{K}_k(x_i,x_i)}{\ell_-}-\sum_{j:y_j=-1}\frac{\widetilde{K}_k(x_i,x_j)}{\ell_-^2}\right)\bigg]^{-1}.
\end{multline*}

\section{Experiments} \label{sec05}

In this section, the behavior of linear and non-linear RLS2 on several learning problems is analyzed. In subsection \ref{sec05.1}, an illustrative analysis of linear RLS2 is proposed, whose goal is to study the feature selection capabilities and the dependence on the regularization parameter of the algorithm in simple experimental settings. RLS2 with non-linear kernels is analyzed in subsection \ref{sec05.2}, where an extensive benchmark on several regression and classification problems from UCI repository is carried out. Finally, multi-class classification of microarray data is considered in subsection \ref{sec05.3}.

Computations are carried out in a Matlab environment and the sub-problem (Problem \ref{PBM06}) is solved using an SMO-like (Sequential Minimal Optimization) algorithm \cite{Platt98}. Current implementation features conjugate gradient to solve linear systems and a sophisticated variable shrinking technique to reduce gradient computations. The stopping criterion for Algorithm \ref{ALG1} used in all the experiments is the following test on the normalized residual of linear system (\ref{E09}):
\[
\|\left(R+\lambda I\right)c-y\| \leq \delta \|y \|.
\]
\noindent The choice $\delta = 10^{-2}$ turns out to be sufficient to make all the coefficients stabilize to a good approximation of their final values. A full discussion of optimization details is outside the scope of the paper. All the experiments have been run on a Core 2 Duo T7700 2.4 GHz, 800 MHz FSB, 4 MB L2 cache, 2 GB RAM.

\subsection{Linear RLS2: illustrative experiments} \label{sec05.1}

In this subsection, we perform two experiments to analyze the behavior of linear RLS2. In the first experiment, a synthetic dataset is used to investigate the ability of linear RLS2 to perform feature selection. The dependence of generalization performances of RLS2 and other learning algorithms on the training set size is analyzed by means of learning curves. The goal of the second experiment is to illustrate the qualitative dependence of coefficients on the regularization parameter and give an idea of the predictive potentiality of the algorithm.

\paragraph{Experiment 1 (\textsc{Binary strings} data)}

\begin{figure}
 \centerline{\includegraphics[width=0.8\linewidth]{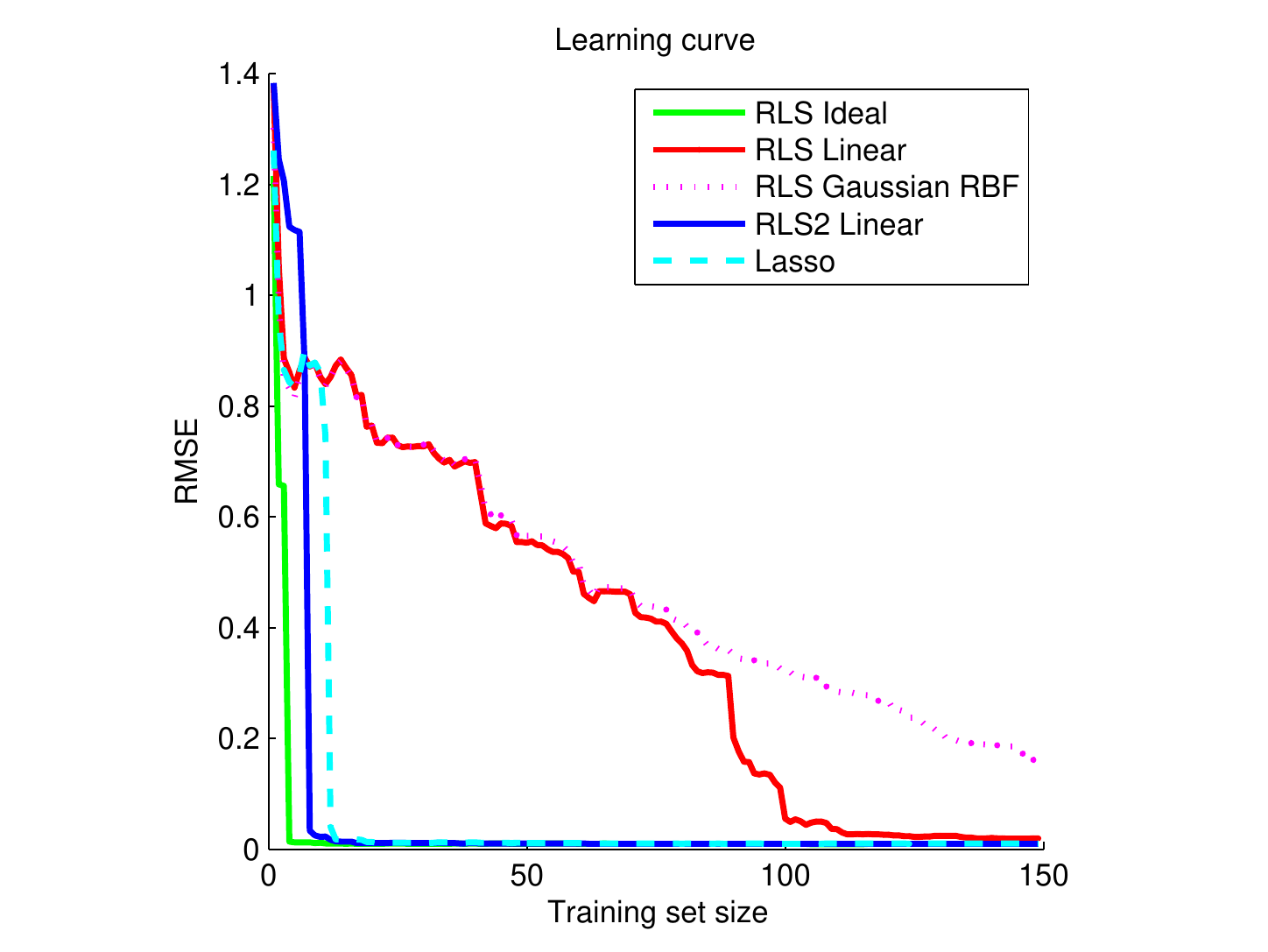} }
 \centerline{\includegraphics[width=0.8\linewidth]{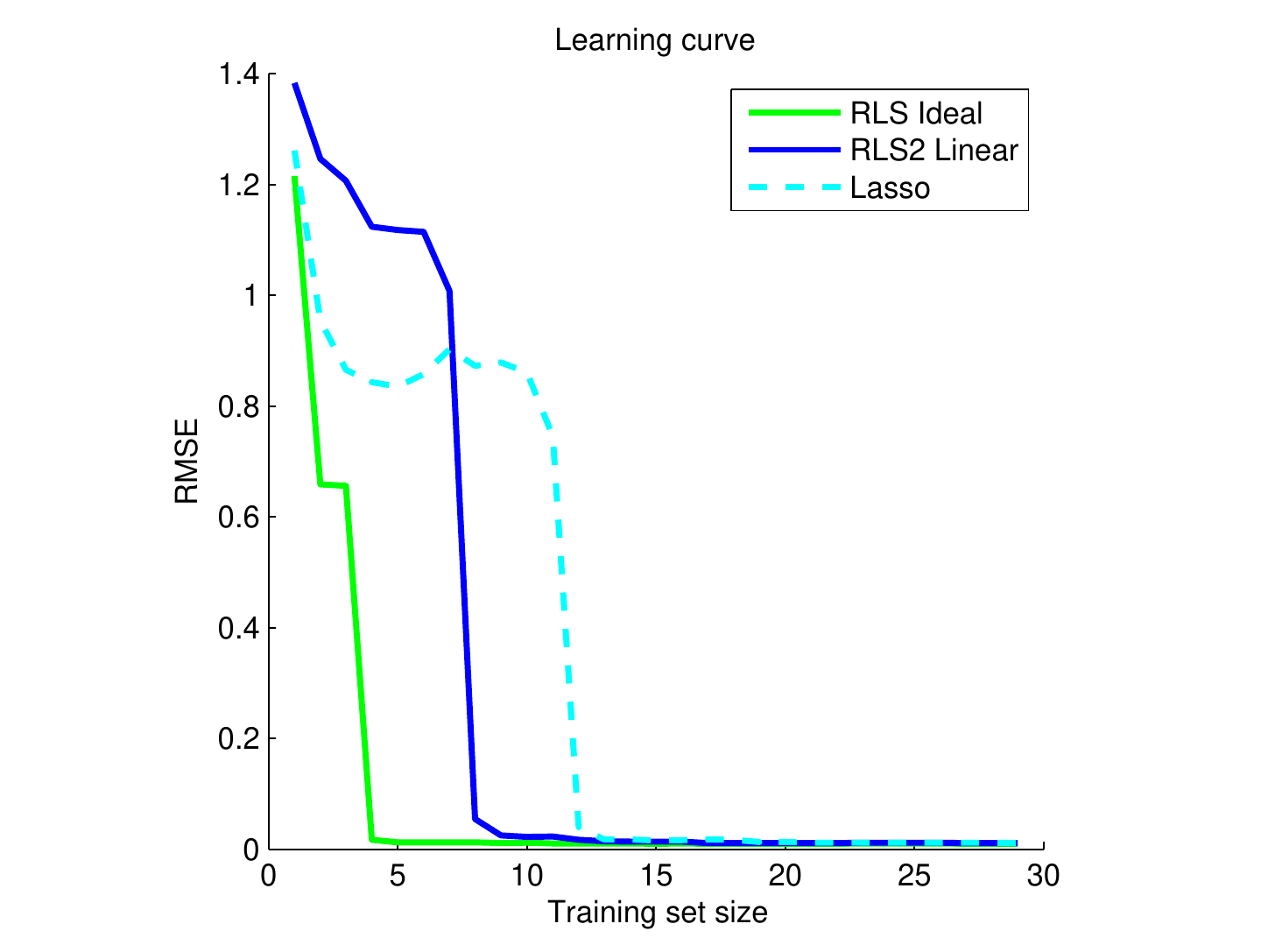} }
 \caption{\textsc{Binary strings} data: lower bounds of RMSE learning curves. The top plot shows test RMSE for training set sizes between 1 and 150 with five different methods: RLS with ideal kernel (see details in the text), RLS with linear kernel (ridge regression), RLS with Gaussian RBF kernel, linear RLS2 and Lasso. The bottom plot is the “zoomed” version of the top plot for training set sizes between 1 and 30, for RLS with ideal kernel, linear RLS2 and Lasso.}\label{FIG11}
\end{figure}

In the first experiment, a synthetic \textsc{Binary strings} dataset has been generated: 250 random binary strings $x_i \in \{0, 1\}^{100}$ are obtained by independently sampling each bit from a Bernoulli distribution with $p = 0.5$. Then, the outputs have been generated as
\[
y_i = x_i^1+x_i^2+x_i^3+\epsilon_i,
\]
\noindent where $\epsilon_i \sim N(0,\sigma^2)$ are small independent Gaussian noises with zero mean and standard deviation $\sigma = 0.01$. In this way, the outputs only depend on the first three bits of the input binary string. The dataset has been divided into a training set of 150 input output pairs and a test set containing the remaining 100 data pairs. We compare the RMSE (root mean squared error) learning curves obtained by varying the training set size using five different methods:
\begin{enumerate}
  \item RLS (regularized least squares) with “ideal” kernel:
    \begin{equation}\label{E19}
    K(x_1,x_2) = x_1^1x_2^1+x_1^2x_2^2+x_1^3x_2^3.
    \end{equation}
  \item RLS with linear kernel (\ref{E13}) (ridge regression).
  \item RLS with Gaussian RBF kernel
    \[
    K(x_1,x_2) = \textrm{exp}\left(-0.01\frac{\|x_1-x_2\|^2}{2}\right).
    \]
  \item RLS2 with linear basis kernels (\ref{E12}) and scaling (\ref{E21}).
  \item Lasso regression.
\end{enumerate}

\noindent  The goal here is to assess the overall quality of regularization paths associated with different regularization algorithms, independently of model selection procedures. To this end, we compute the RMSE on the test data as a function of the training set size and evaluate the lower bounds of the learning curves with respect to variation of the regularization parameter. Results are shown in Figure \ref{FIG11}, whose top plot reports the lower bounds of learning curves for all the five algorithms with training set sizes between 1 and 150. Notice that all the five methods are able to learn, asymptotically, the underlying “concept”, up to the precision limit imposed by the noise, but methods that exploits coefficients sparsity are faster to reach the asymptotic error rate. Not surprisingly, the best method is RLS with the “ideal kernel” (\ref{E19}), which incorporates a strong prior knowledge: the dependence of the outputs on the first three bits only. Though knowing in advance the optimal features is not realistic, this method can be used as a reference. The slowest learning curve is that associated to RLS with Gaussian RBF kernel, which only incorporates a notion of smoothness. A good compromise is RLS with linear kernel, which uses the knowledge of linearity of the underlying function, and reaches a good approximation of the asymptotic error rate after seeing about 100 strings. The remaining two methods (Lasso and linear RLS2) incorporate the knowledge of both linearity and sparsity. They are able to learn the underlying concept after seeing only 12 examples, despite the presence of the noise. Since after the 12-th example Lasso and linear RLS2 are basically equivalent, is it interesting to see what happen for very small sample sizes. The bottom plot of Figure \ref{FIG11} is a zoomed version of the top plot with training set sizes between 1 and 30, showing only the learning curves for the three methods that impose sparsity. Until the 8-th example, the Lasso learning curve stays lower than the RLS2 learning curve. After the 8-th example, the RLS2 learning curve stays uniformly lower than the Lasso, indicating an high efficiency in learning noisy sparse linear combinations. Since the multiple kernel learning interpretation of RLS2 suggests that the algorithm is being learning the “ideal” kernel (\ref{E19}) simultaneously with the predictor, it might be interesting to analyze the asymptotic values of kernel coefficients $d_i$.  Indeed, after the first 12 training examples, RLS2 sets to zero all the coefficients $d_i$ except the first three, which are approximately equal to 1/3.

\paragraph{Experiment 2 (\textsc{Prostate Cancer} data)}

\begin{figure}
 \centerline{\includegraphics[width=1\linewidth]{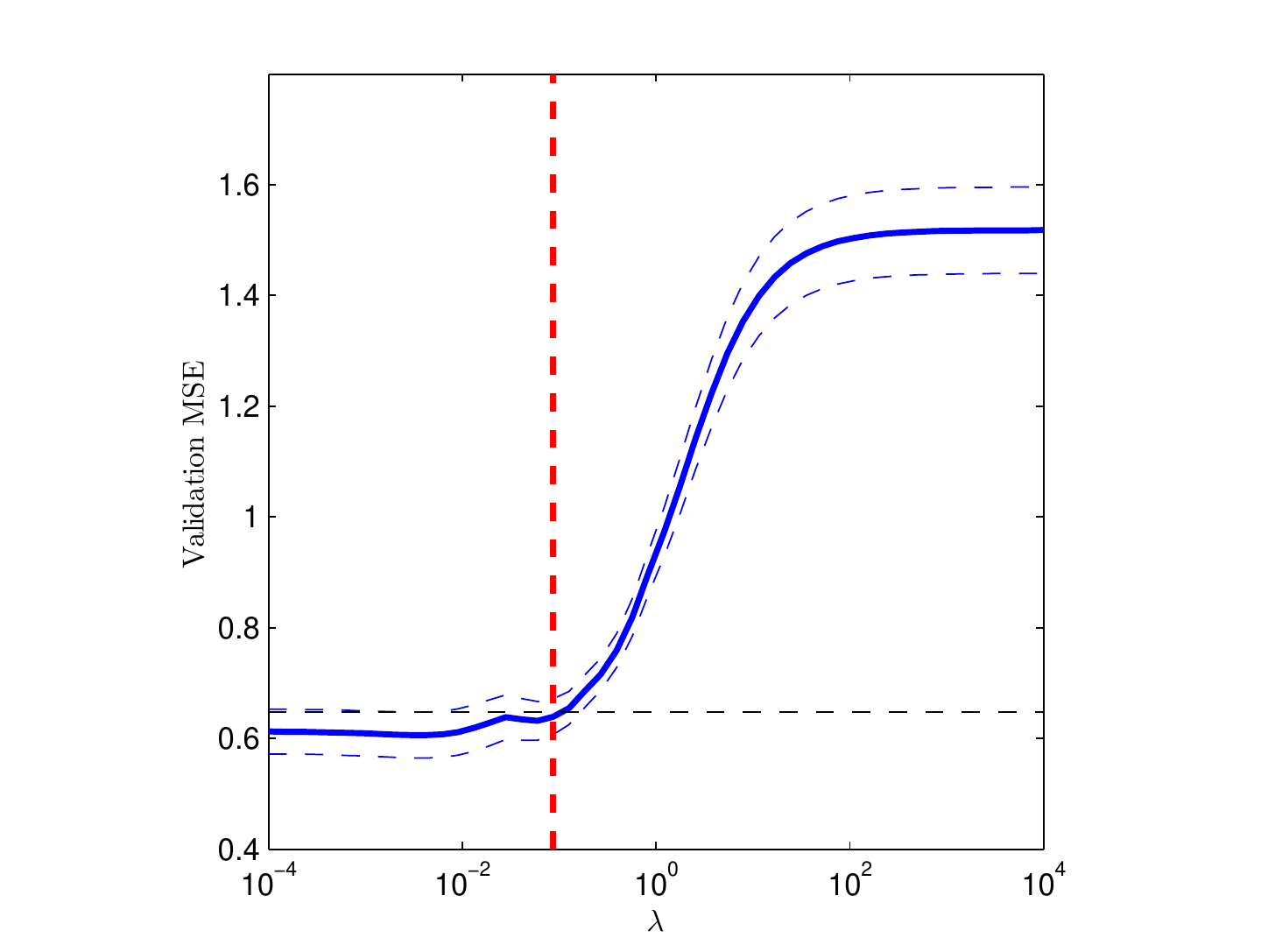} }
 \caption{\textsc{Prostate Cancer} data: 10-fold cross-validation prediction error curves and their standard errors bands estimated for linear RLS2. Model complexity increases from the right to the left. The vertical line corresponds to the least complex model within one standard error of the best.}\label{FIG10}
\end{figure}
\begin{figure}
\centerline{\includegraphics[width=1\linewidth]{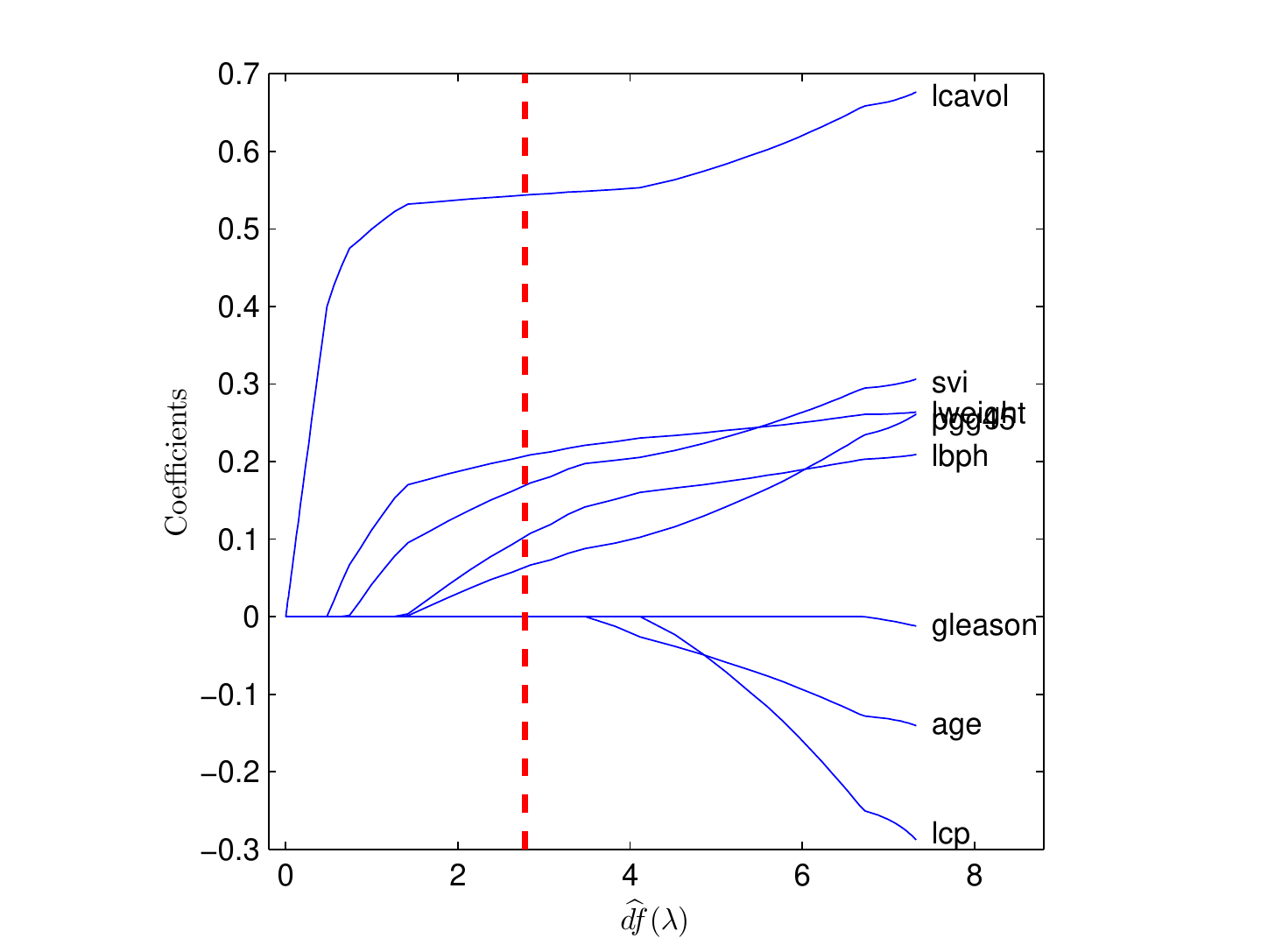}}
 \caption{\textsc{Prostate Cancer} data: profiles of RLS2 coefficients with respect to a continuous variation of the regularization parameter. Coefficients are plotted versus $\widehat{df}(\lambda)$, the approximate degrees of freedom. The vertical line corresponds to the value of $\lambda$ chosen in the validation phase.}\label{FIG11}
\end{figure}

\begin{table}
  \caption{\textsc{Prostate Cancer} data: comparison of RLS2 with other subset selection and shrinkage methods. Estimated coefficients, test error and their standard error are reported. Results for methods other than RLS2 are taken from \cite{Hastie08}. Blank entries corresponds to variables not selected.}
  \centering
  \begin{tabular}{|lccccccc|}
  \hline
  Term                  & RLS2 & Best subset   & LS    & Ridge & Lasso & PCR   & PLS   \\
  \hline
  \textsc{Intercept}    & 2.452 & 2.477         & 2.465 & 2.452 & 2.468 & 2.497 & 2.452 \\
  \textsc{lcavol}       & 0.544 & 0.740         & 0.680 & 0.420 & 0.533 & 0.543 & 0.419 \\
  \textsc{lweight}      & 0.207 & 0.316         & 0.263 & 0.238 & 0.169 & 0.289 & 0.344 \\
  \textsc{age}          &       &               &-0.141 &-0.046 &       &-0.152 &-0.026 \\
  \textsc{lbph}         & 0.104 &               & 0.210 & 0.162 & 0.002 & 0.214 & 0.220 \\
  \textsc{svi}          & 0.170 &               & 0.305 & 0.227 & 0.094 & 0.315 & 0.243 \\
  \textsc{lcp}          &       &               &-0.288 & 0.000 &       &-0.051 & 0.079 \\
  \textsc{gleason}      &       &               &-0.021 & 0.040 &       & 0.232 & 0.011 \\
  \textsc{pgg45}        & 0.064 &               & 0.267 & 0.133 &       &-0.056 & 0.084 \\
  \hline
  Test error            & 0.454 & 0.492         & 0.521 & 0.492 & 0.479 & 0.449 & 0.528 \\
  Std error             & 0.152 & 0.143         & 0.179 & 0.165 & 0.164 & 0.105 & 0.152 \\
  \hline
  \end{tabular}
  \label{TAB10}
\end{table}

Linear RLS2 is applied to the \textsc{Prostate Cancer} dataset, a regression problem whose goal is to predict the level of prostate-specific antigen on the basis of a number of clinical measures in men who were about to receive a radical prostatectomy \cite{Stamey89}. These data are used in the textbook \cite{Hastie08} to compare different feature selection and shrinkage methods, and have been obtained from the web site \url{http://www-stat.stanford.edu/ElemStatLearn/}. Data have been preprocessed by normalizing all the inputs to zero mean and unit standard deviation. The dataset is divided into a training set of 67 examples and a test set of 30 examples. To choose the regularization parameter, the 10-fold cross-validation score has been computed for different values of $\lambda$ in the interval $\left[10^{-4}, 10^{4}\right]$ on a logarithmic scale. The scaling coefficients $s_i$ are chosen as in (\ref{E21}), thus normalizing each training feature to have unit norm. An intercept term equal to the average of training outputs has been subtracted to the outputs before estimating the other coefficients. For each of the dataset splits, the MSE (mean squared error) has been computed on the validation data. Figure \ref{FIG10} reports average and standard error bands for validation MSE along a regularization path. Following \cite{Hastie08}, we pick the value of $\lambda$ corresponding to the least complex model within one standard error of the best validation score.

In a second phase, the whole training set (67 examples) is used to compute the RLS2 solution with different values of $\lambda$. Figure \ref{FIG11} reports the profile of RLS2 coefficients $a_j$, see equation (\ref{E17}), along the whole regularization path as a function of the degrees of freedom defined as in (\ref{E14}). RLS2 does a continuous feature selection that may resemble that of the Lasso. However, the dependence of coefficients on the regularization parameter is rather complex and the profile in Figure \ref{FIG10} is not piecewise linear. In correspondence with the value of $\lambda$ chosen in the validation phase, RLS2 selects 5 input variables out of 8. Table \ref{TAB10} reports the value of coefficients estimated by RLS2 together with the test error and his standard error. For comparison, Table \ref{TAB10} also reports models and results taken from \cite{Hastie08} associated with LS (Least Squares), Best subset regression, Ridge Regression, Lasso regression, PCR (Principal Component Regression), PLS (Partial Least Squares). The best model on these data is PCR, but RLS2 achieves the second lowest test error by using only 5 variables.

\subsection{Non-linear RLS2: regression and classification benchmark} \label{sec05.2}

\begin{table}
  \caption{Data sets used in the experiments. The first four are regression problems while the last six are classification problems.}
  \centering
  \begin{tabular}{|lcccc|}
  \hline
  Dataset               & Feature standardization       & Examples  & Features  & Kernels   \\
  \hline
  \textsc{Auto-mpg}     & Yes                           & 392       & 7         & 104       \\
  \textsc{Cpu}          & Yes                           & 209       & 8         & 494       \\
  \textsc{Servo}        & No                            & 167       & 4         & 156       \\
  \textsc{Housing}      & Yes                           & 506       & 13        & 182       \\
  \hline
  \hline
  \textsc{Heart}        & Yes                           & 270       & 13        & 182       \\
  \textsc{Liver}        & No                            & 345       & 6         & 91        \\
  \textsc{Pima}         & Yes                           & 768       & 8         & 117       \\
  \textsc{Ionosphere}   & Yes                           & 351       & 33        & 442       \\
  \textsc{Wpbc}         & Yes                           & 194       & 34        & 455       \\
  \textsc{Sonar}        & No                            & 208       & 60        & 793       \\
  \hline
  \end{tabular}
  \label{TAB1}
\end{table}

\begin{figure}
 \centerline{\includegraphics[width=0.4\linewidth]{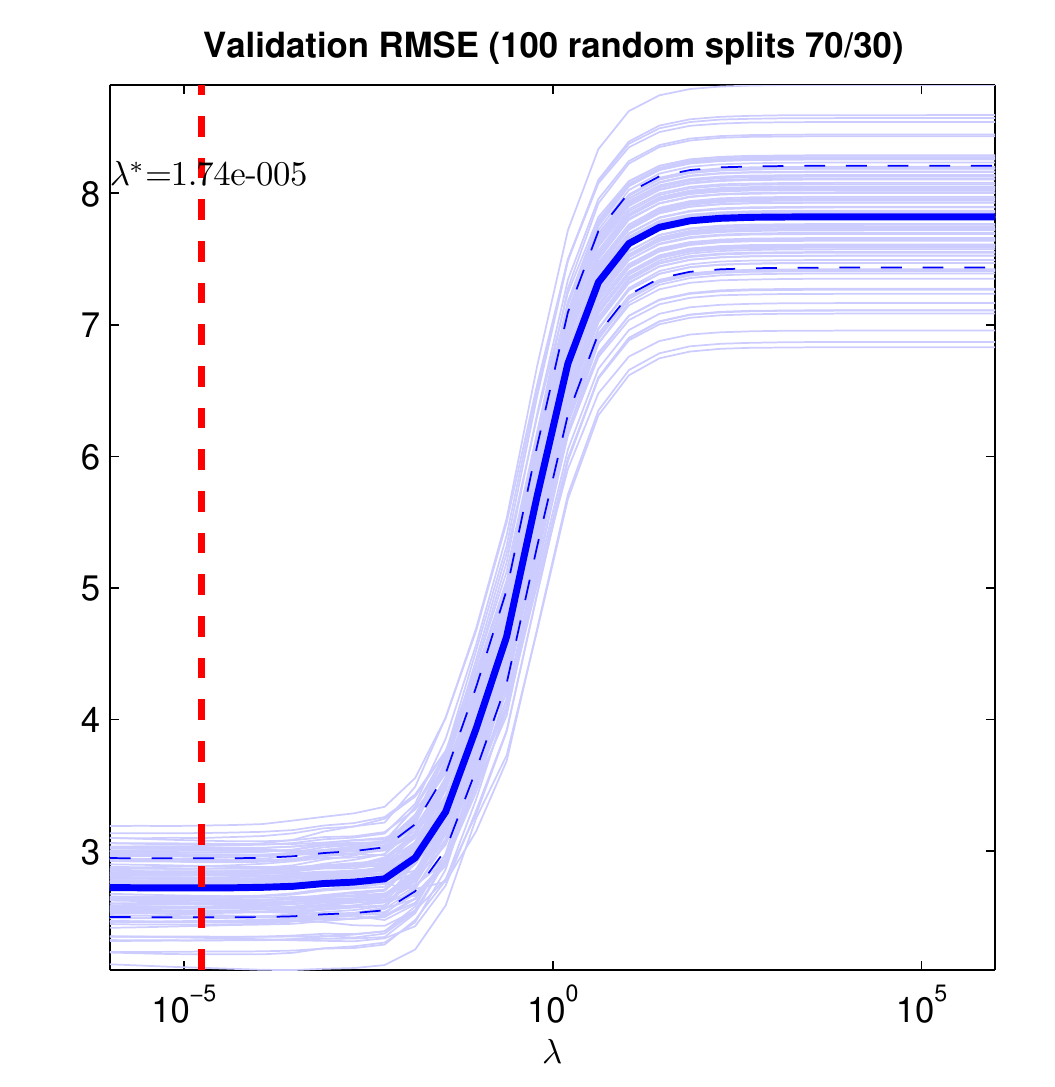}  \includegraphics[width=0.4\linewidth]{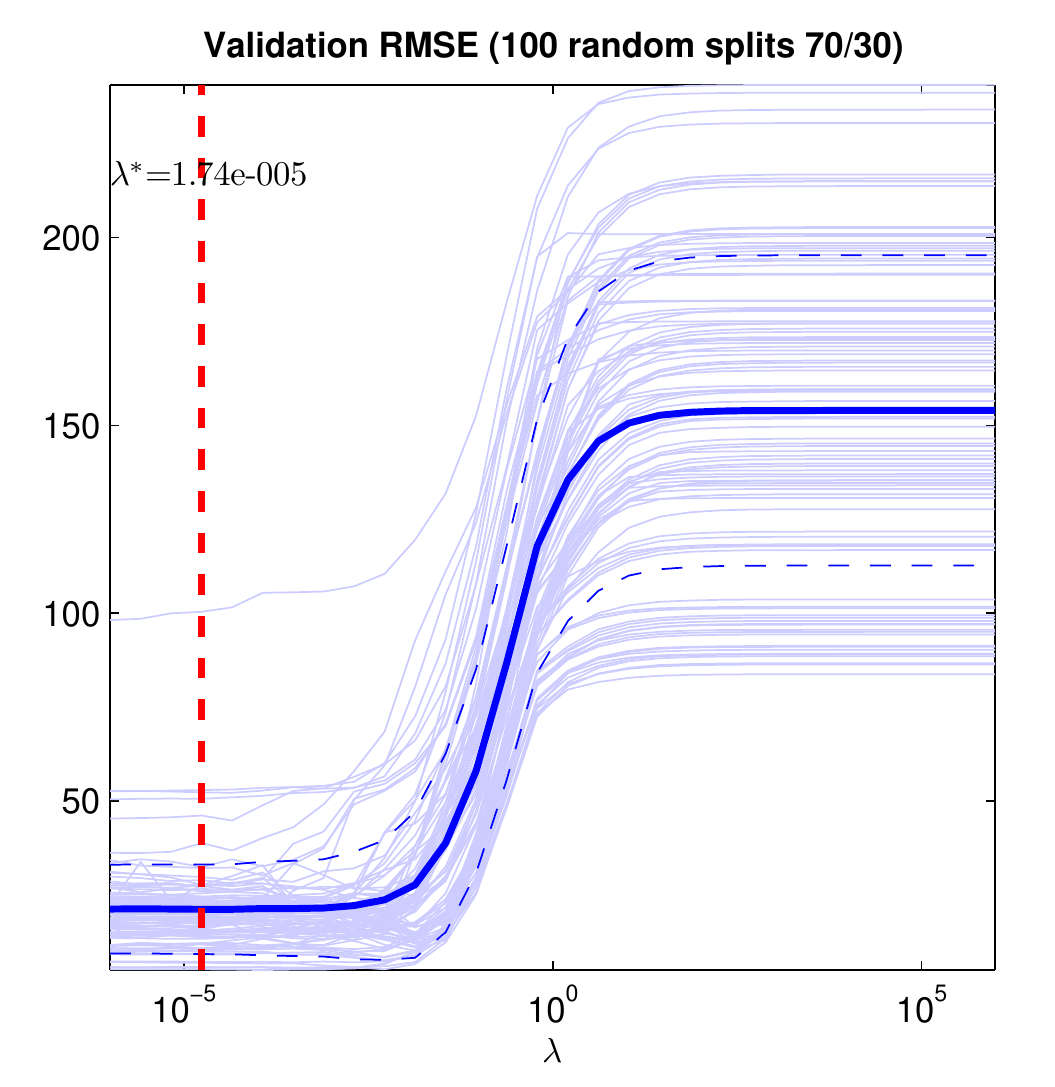}}
 \centerline{\includegraphics[width=0.4\linewidth]{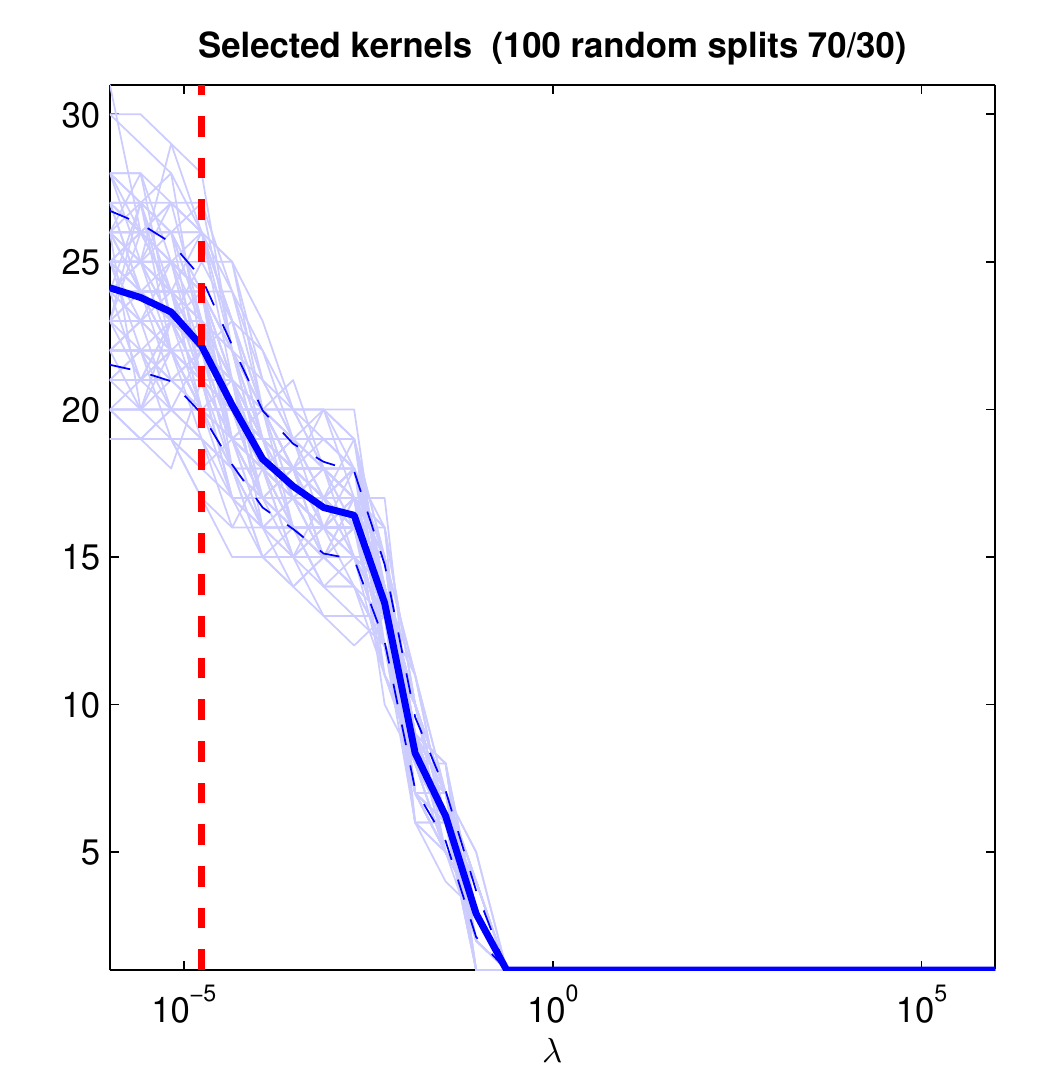}  \includegraphics[width=0.4\linewidth]{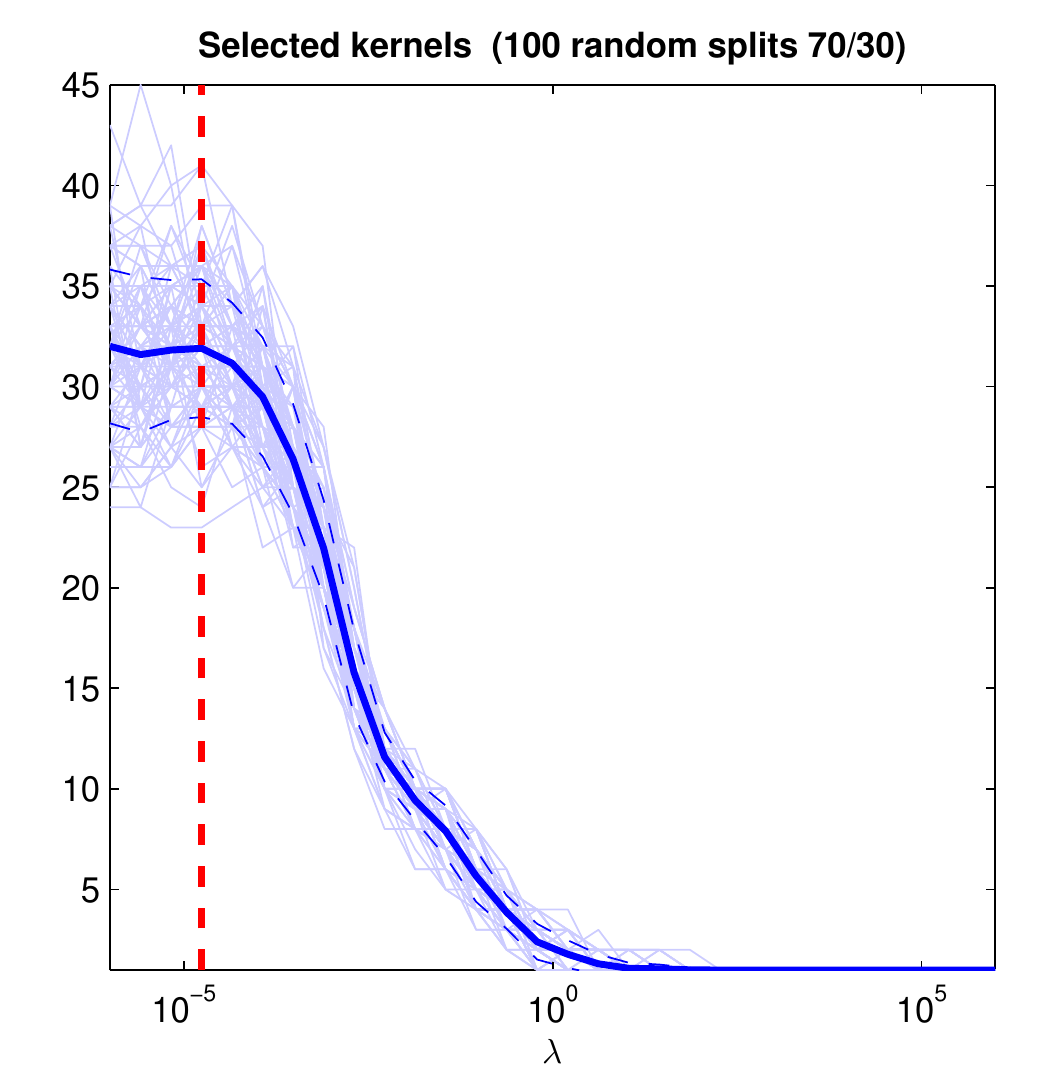}}
 \centerline{\includegraphics[width=0.4\linewidth]{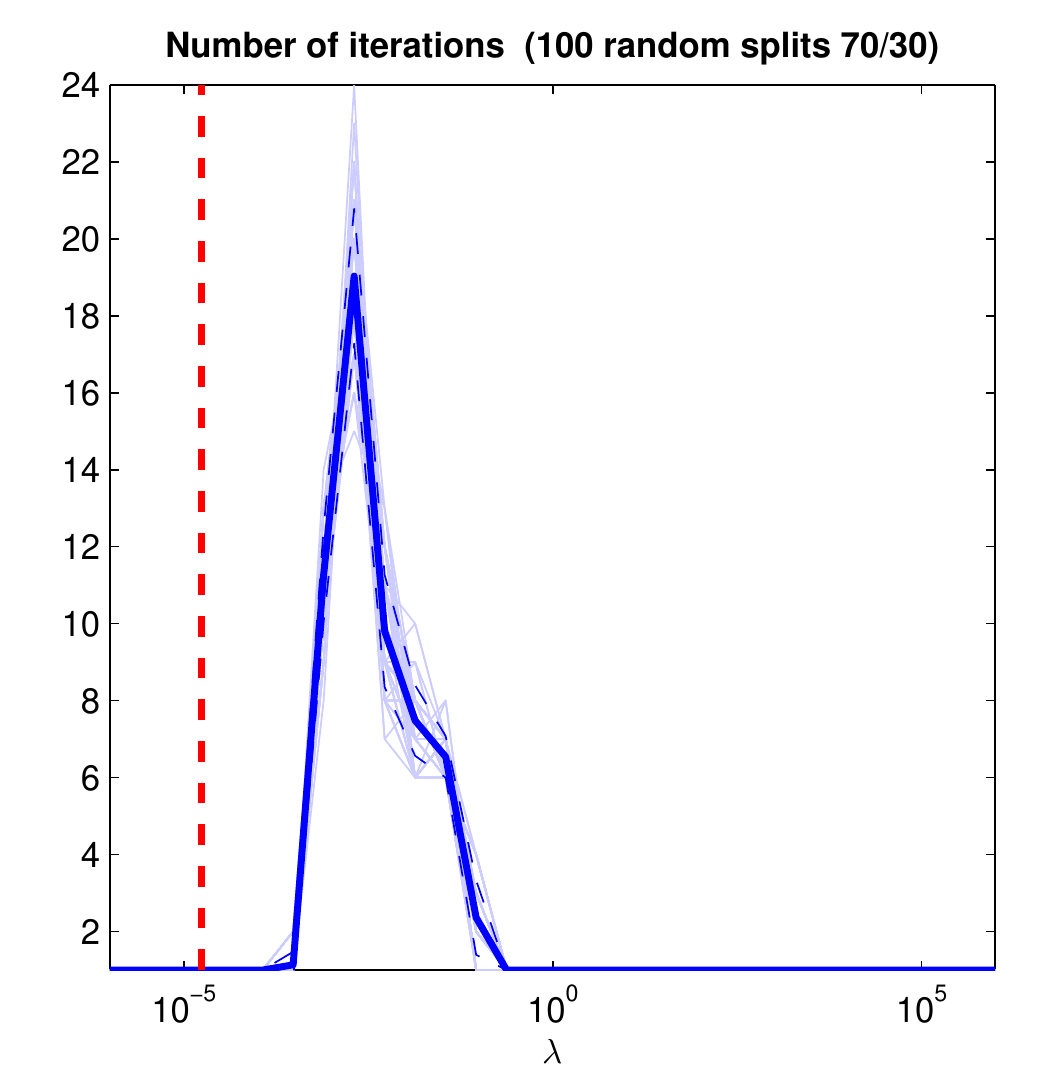}  \includegraphics[width=0.4\linewidth]{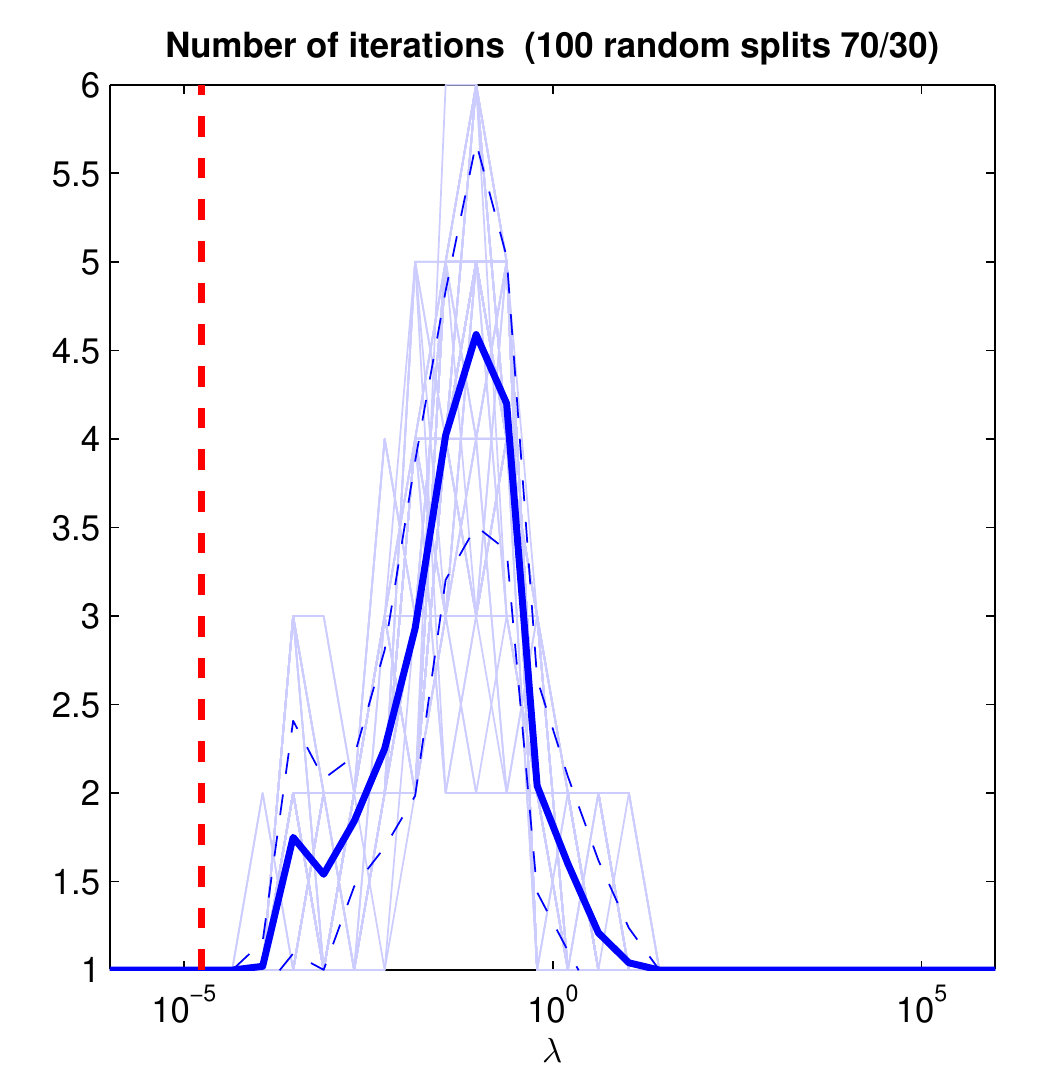}}
  \caption{RLS2 on the \textsc{Auto-mpg} (left) and the \textsc{Cpu} (right) dataset: RMSE on the test data (top), number of selected kernels (center), and number of iterations (bottom) along a regularization path.}\label{FIG01}
\end{figure}

\begin{figure}
 \centerline{\includegraphics[width=0.4\linewidth]{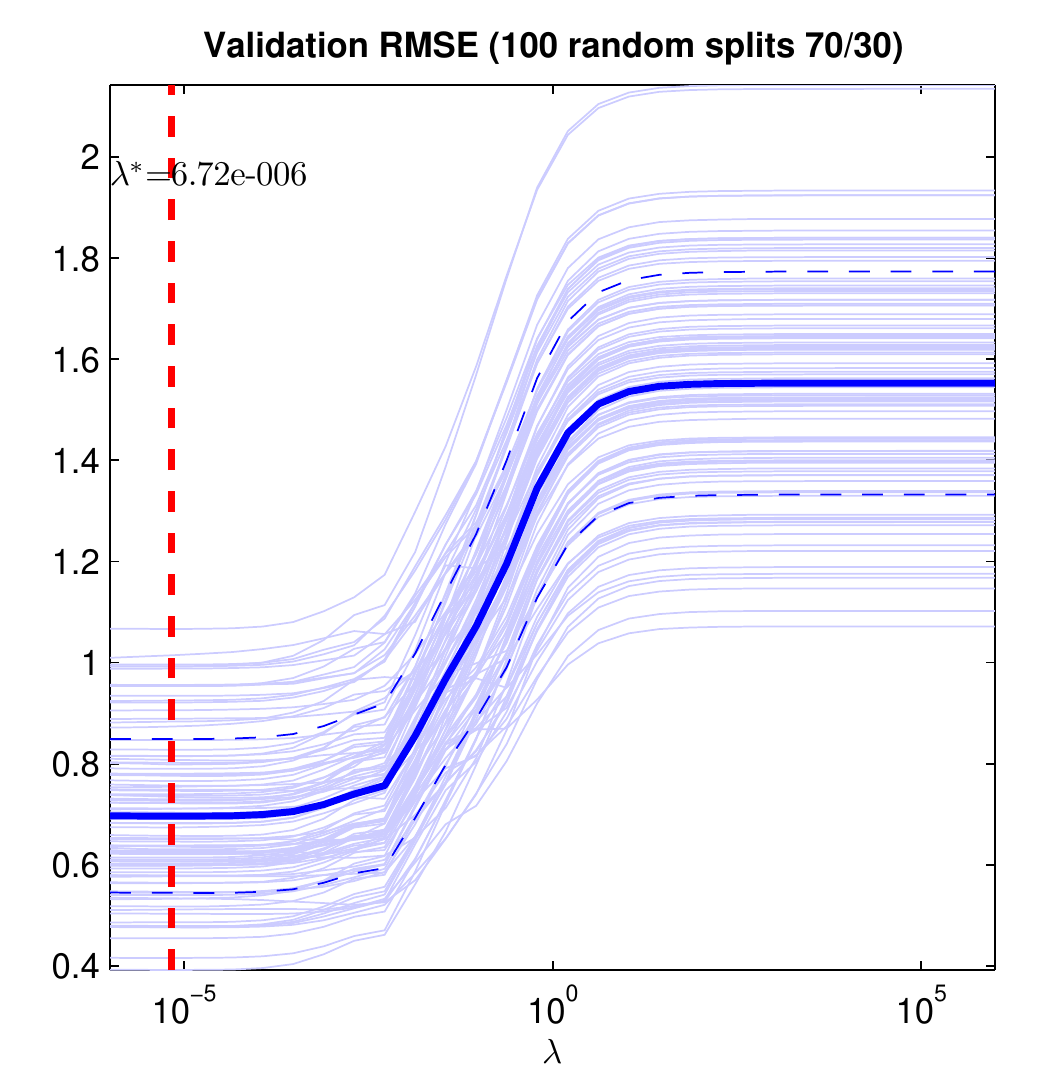}\includegraphics[width=0.4\linewidth]{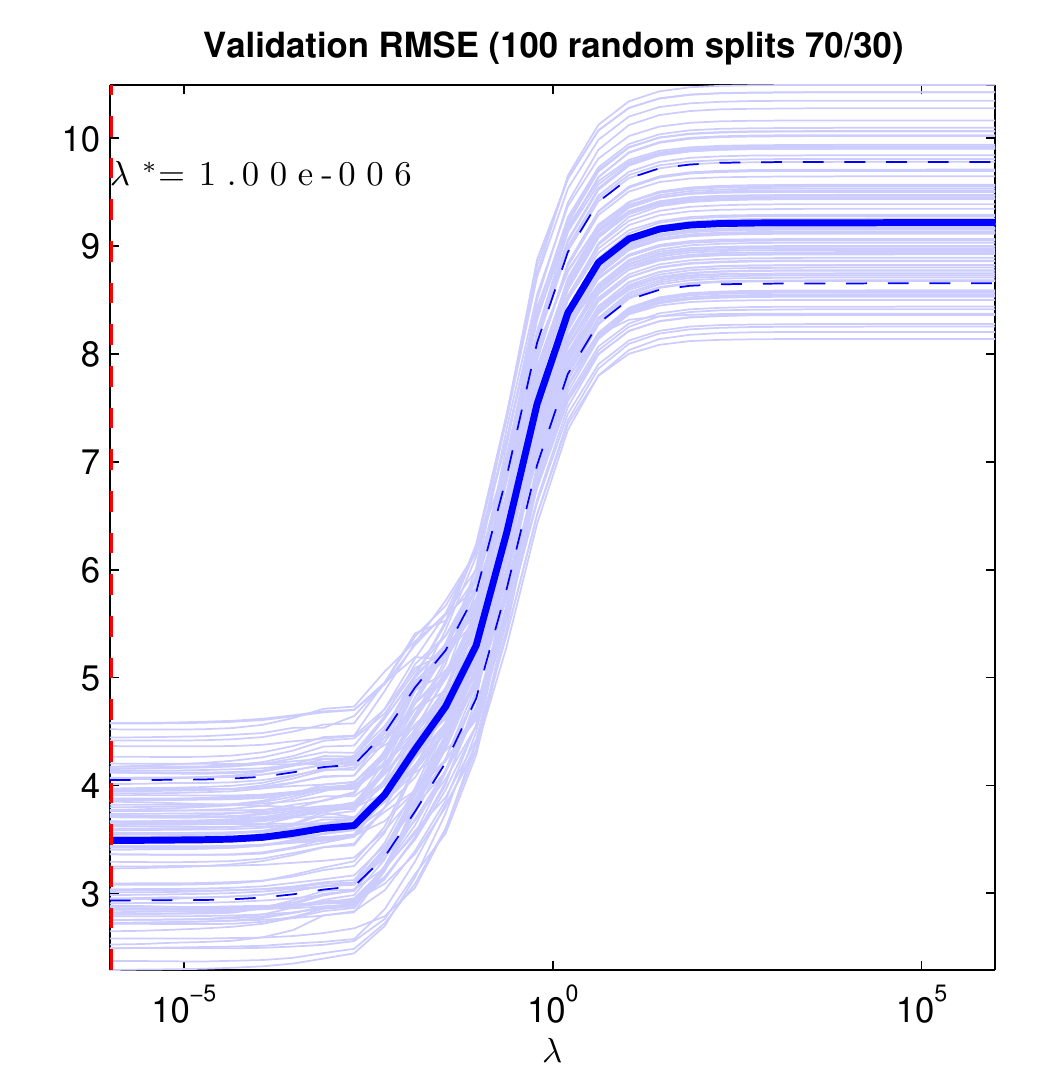}}
 \centerline{\includegraphics[width=0.4\linewidth]{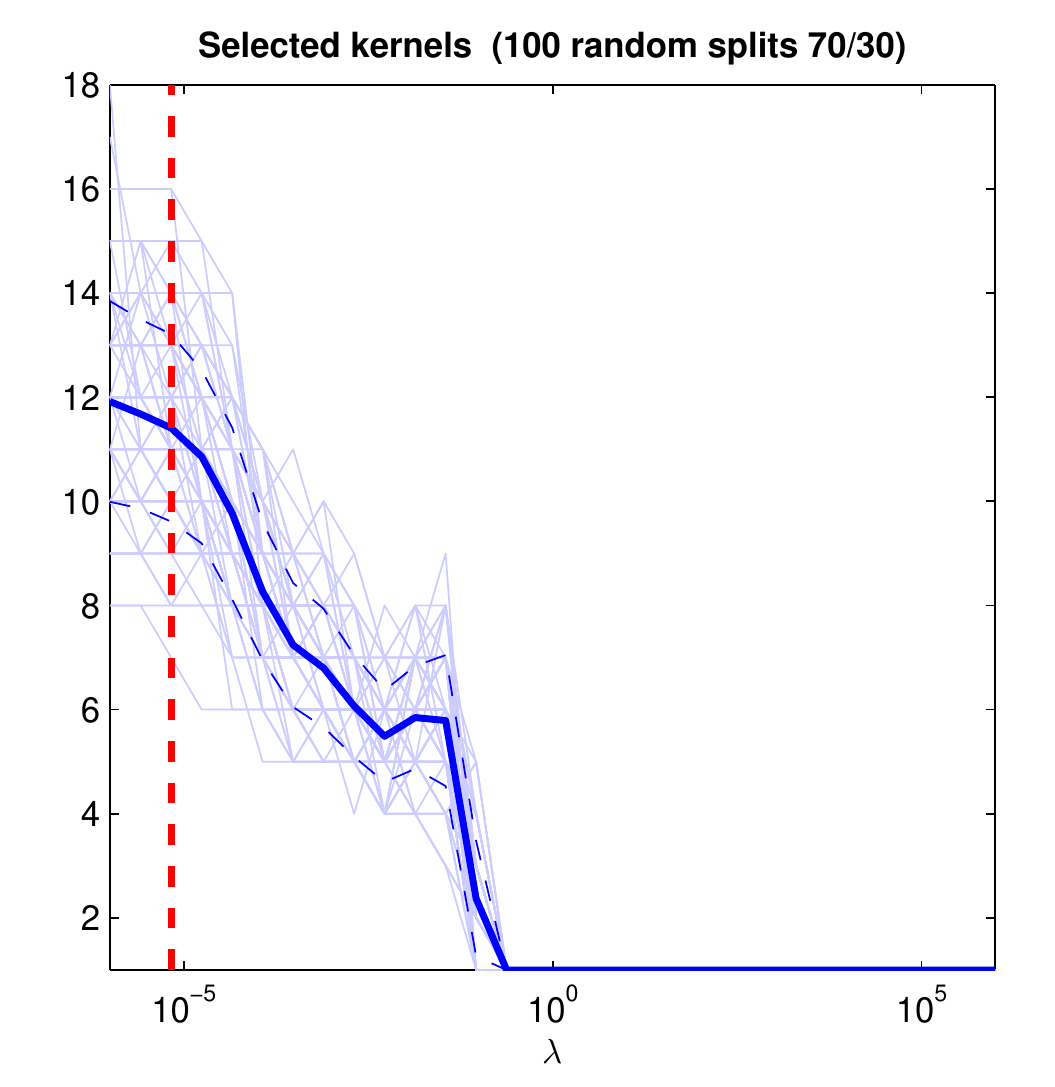}  \includegraphics[width=0.4\linewidth]{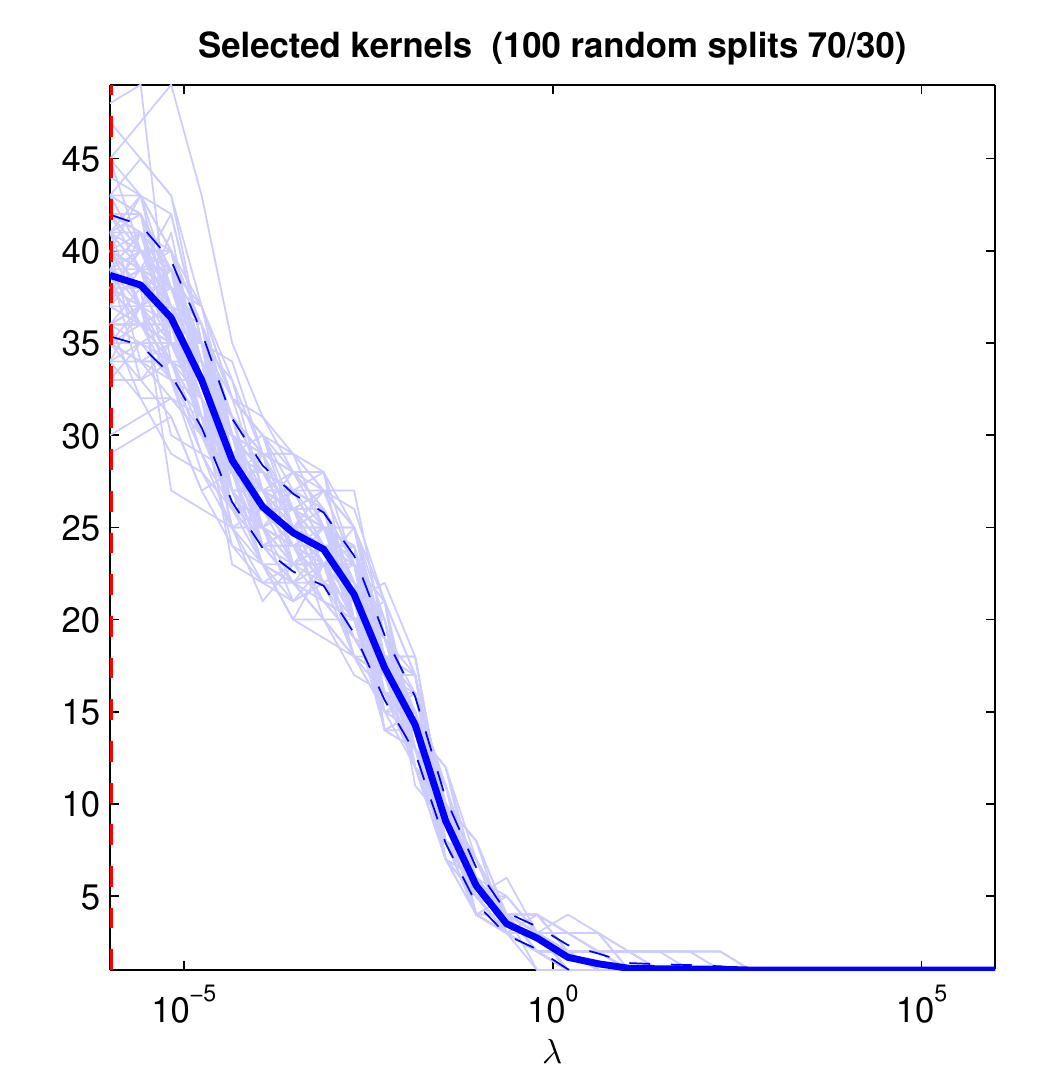}}
 \centerline{\includegraphics[width=0.4\linewidth]{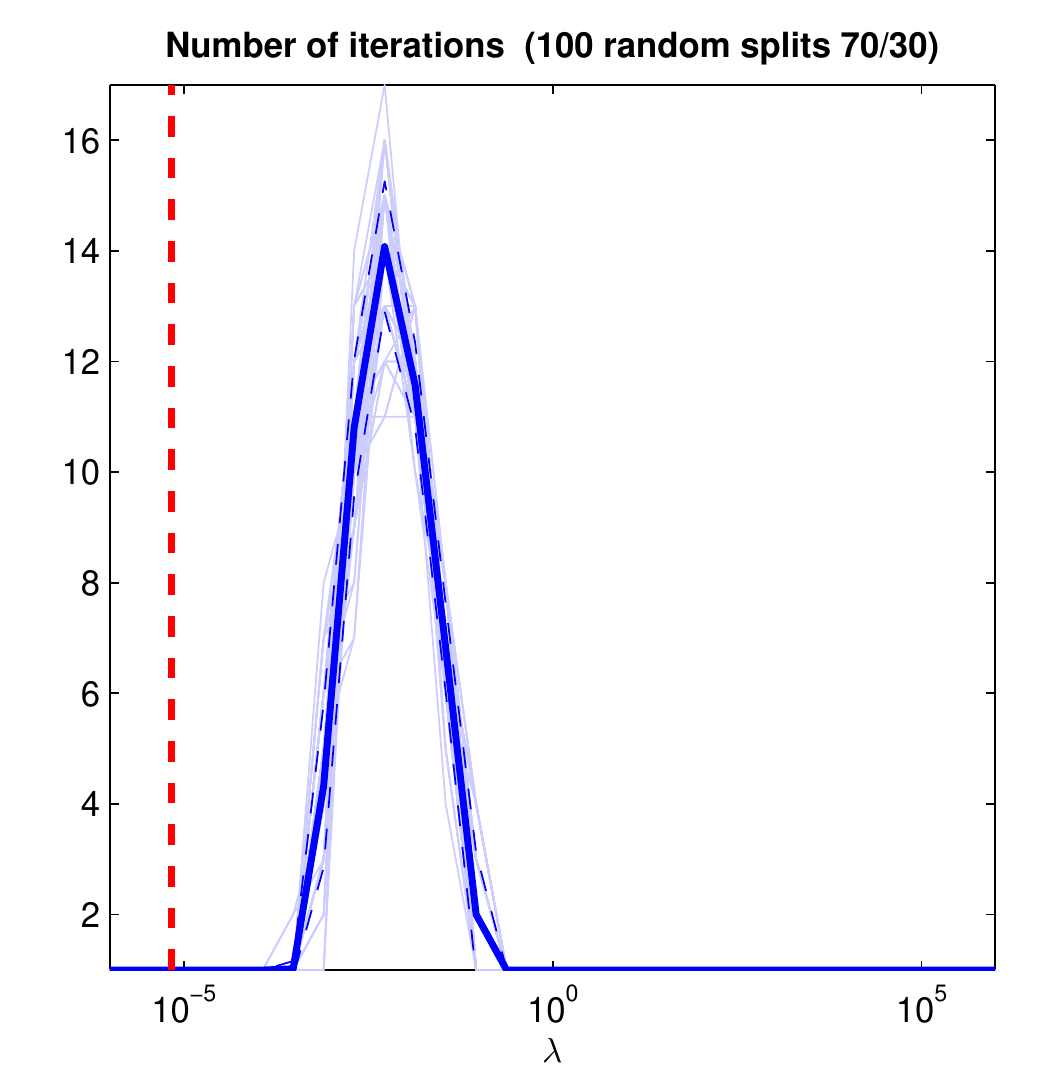}  \includegraphics[width=0.4\linewidth]{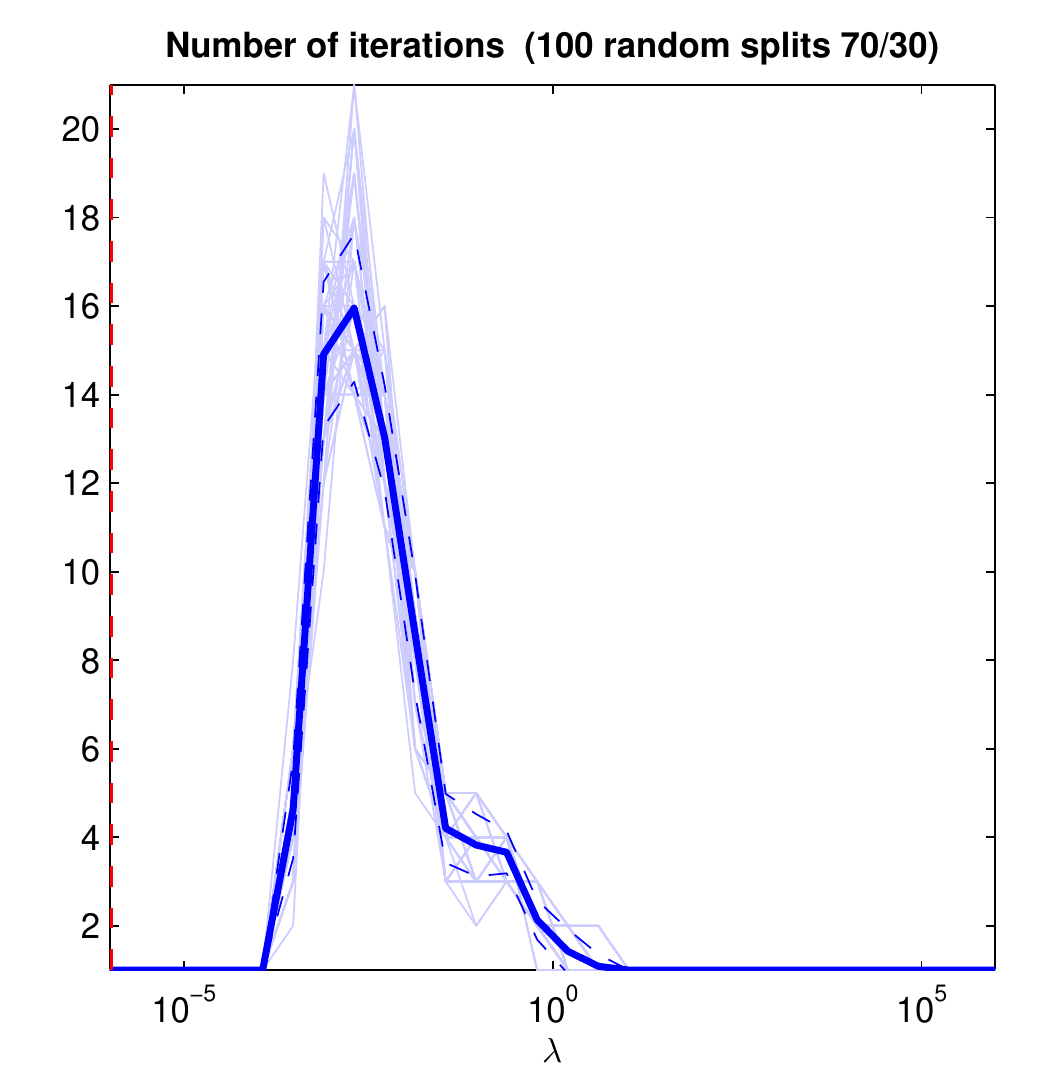}}
  \caption{RLS2 on the \textsc{Servo} (left) and the \textsc{Housing} (right) dataset: RMSE on the test data (top), number of selected kernels (center), and number of iterations (bottom) along a regularization path.}\label{FIG02}
\end{figure}

\begin{figure}
 \centerline{\includegraphics[width=0.4\linewidth]{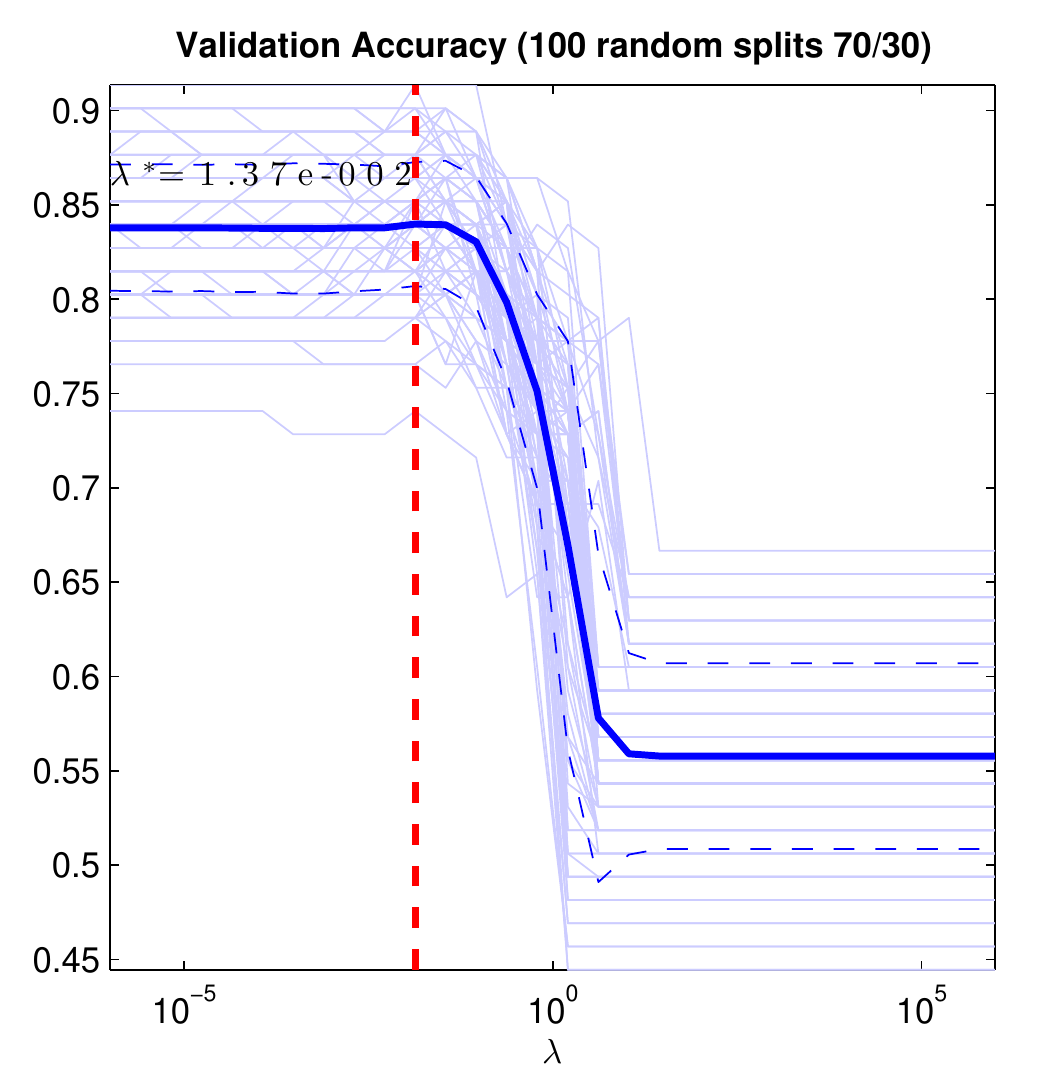}  \includegraphics[width=0.4\linewidth]{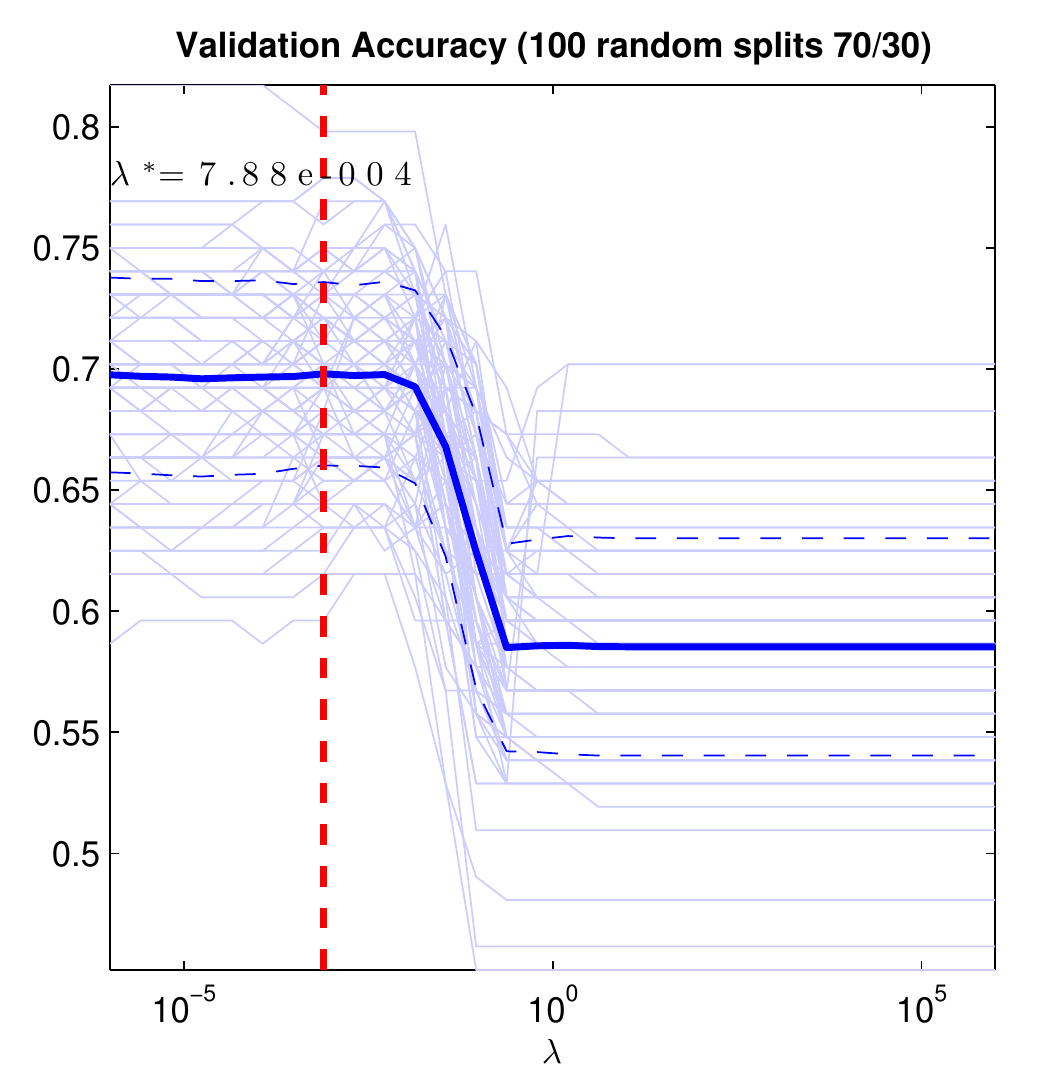}}
 \centerline{\includegraphics[width=0.4\linewidth]{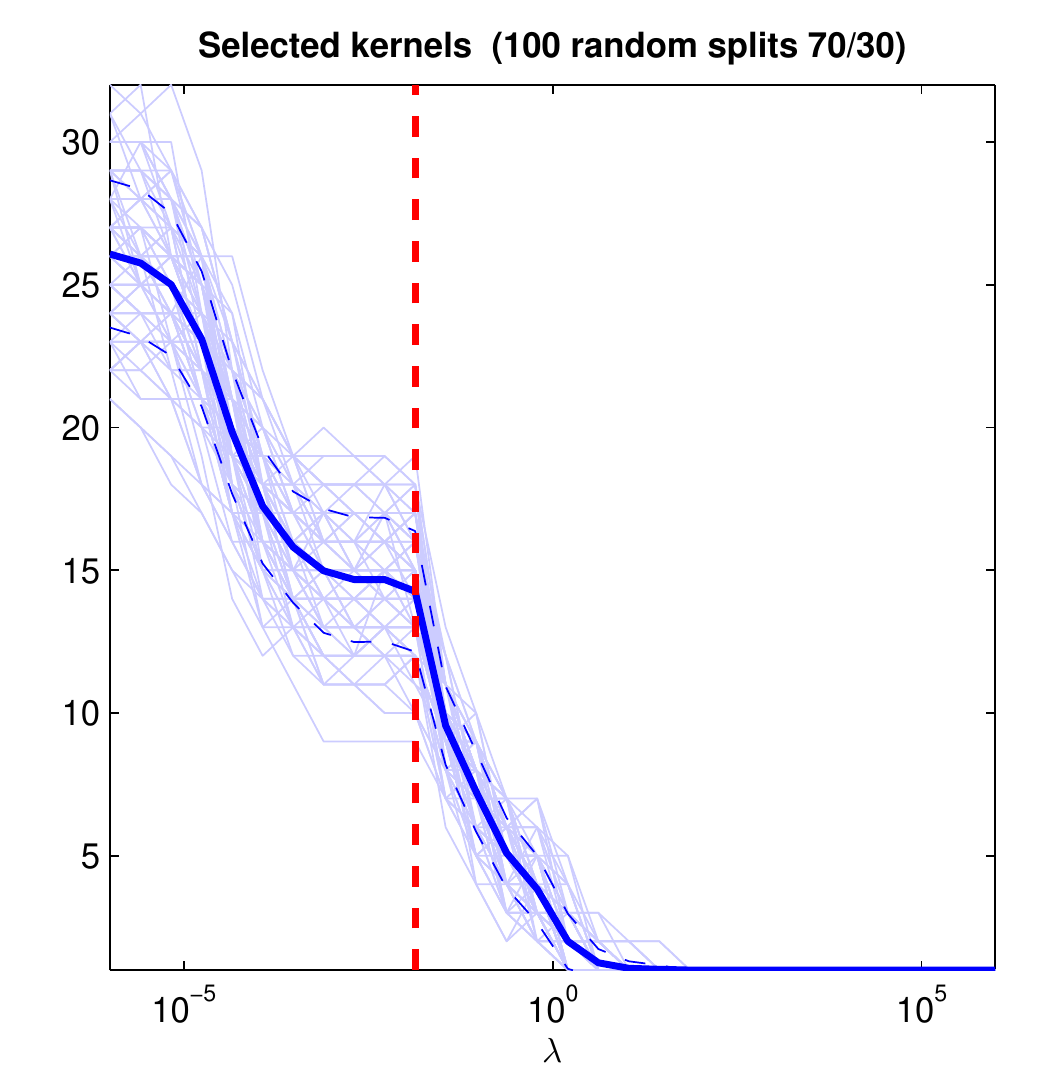}  \includegraphics[width=0.4\linewidth]{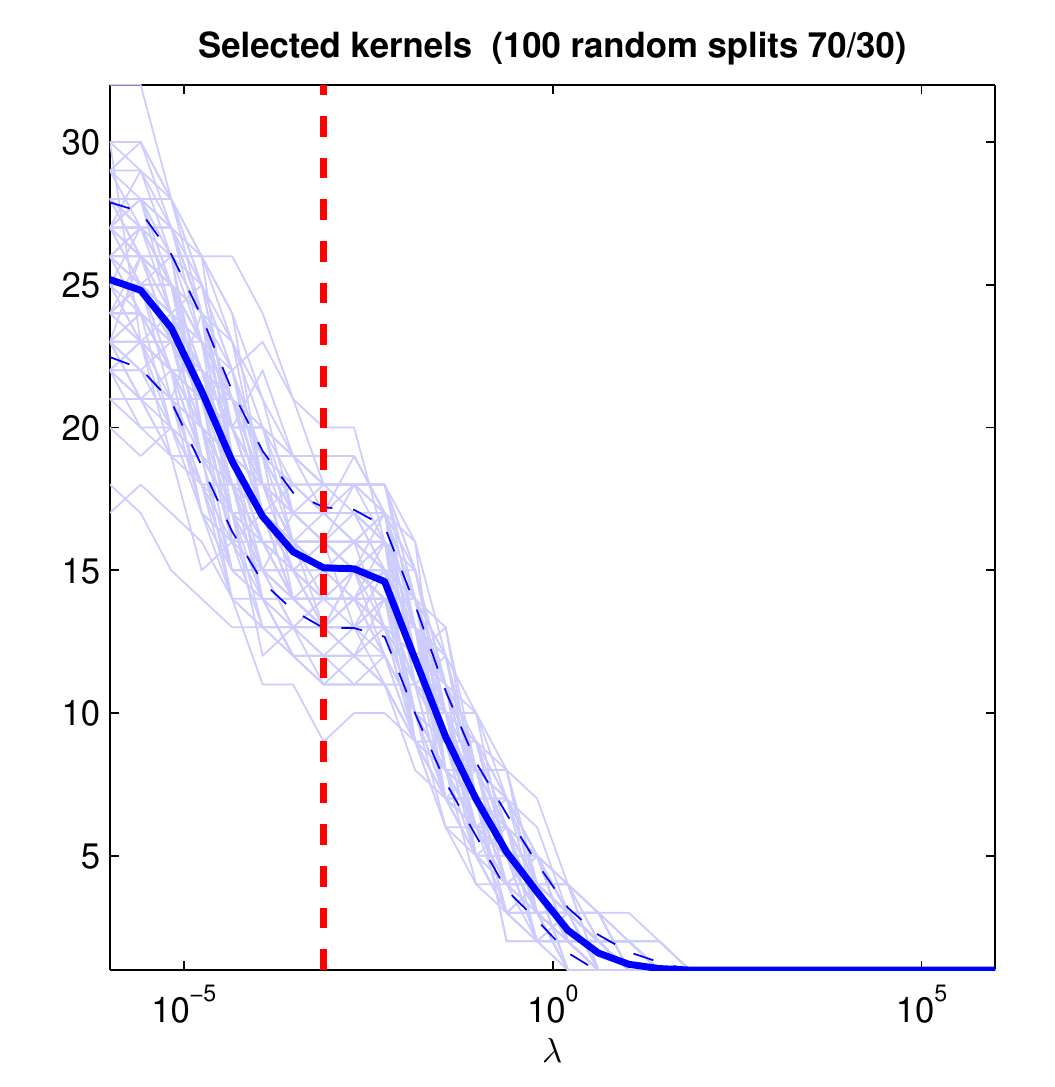}}
 \centerline{\includegraphics[width=0.4\linewidth]{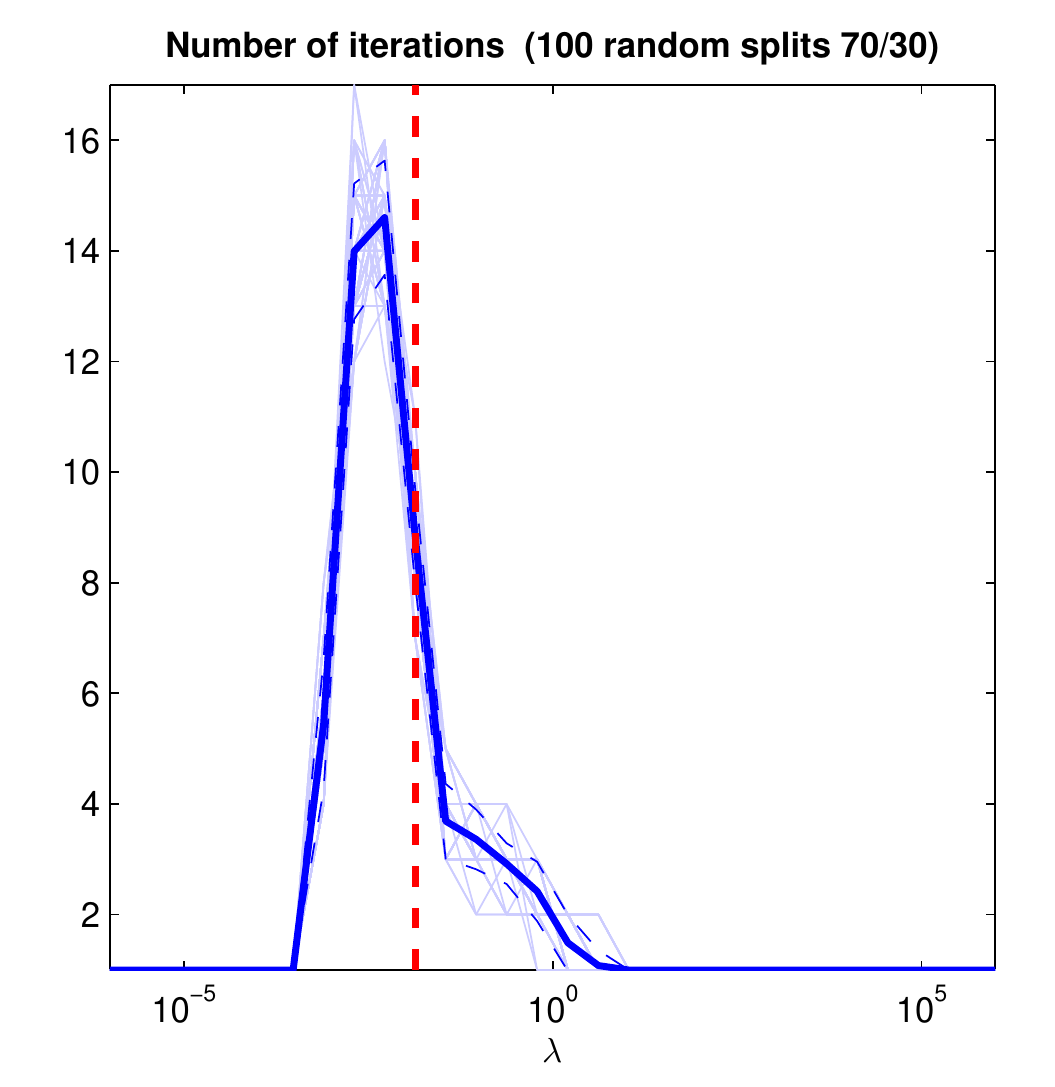}  \includegraphics[width=0.4\linewidth]{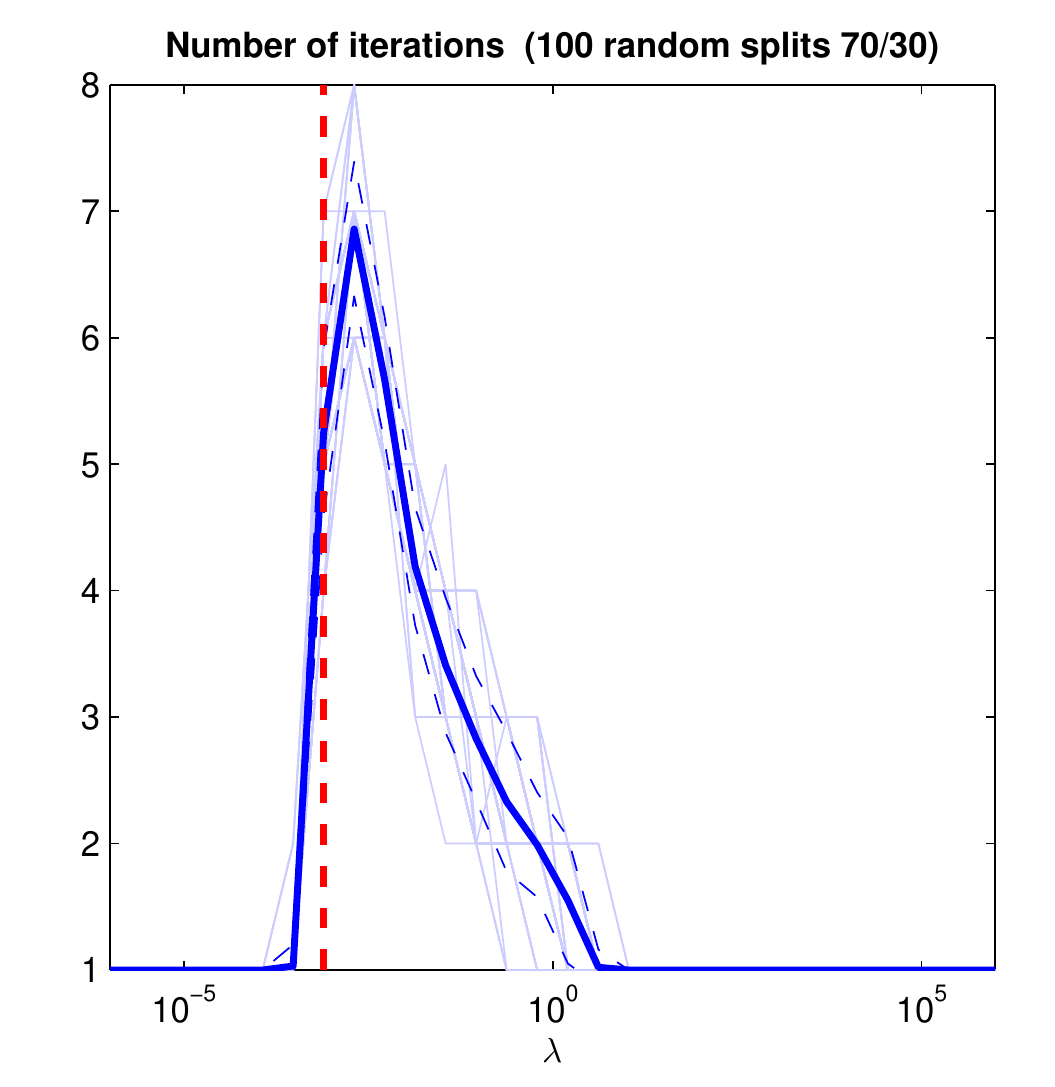}}
  \caption{RLS2 on the \textsc{Heart} (left) and \textsc{Liver} (right) dataset: classification accuracy on the test data (top), number of selected kernels (center), and number of iterations (bottom) along a regularization path.}\label{FIG03}
\end{figure}

\begin{figure}
 \centerline{\includegraphics[width=0.4\linewidth]{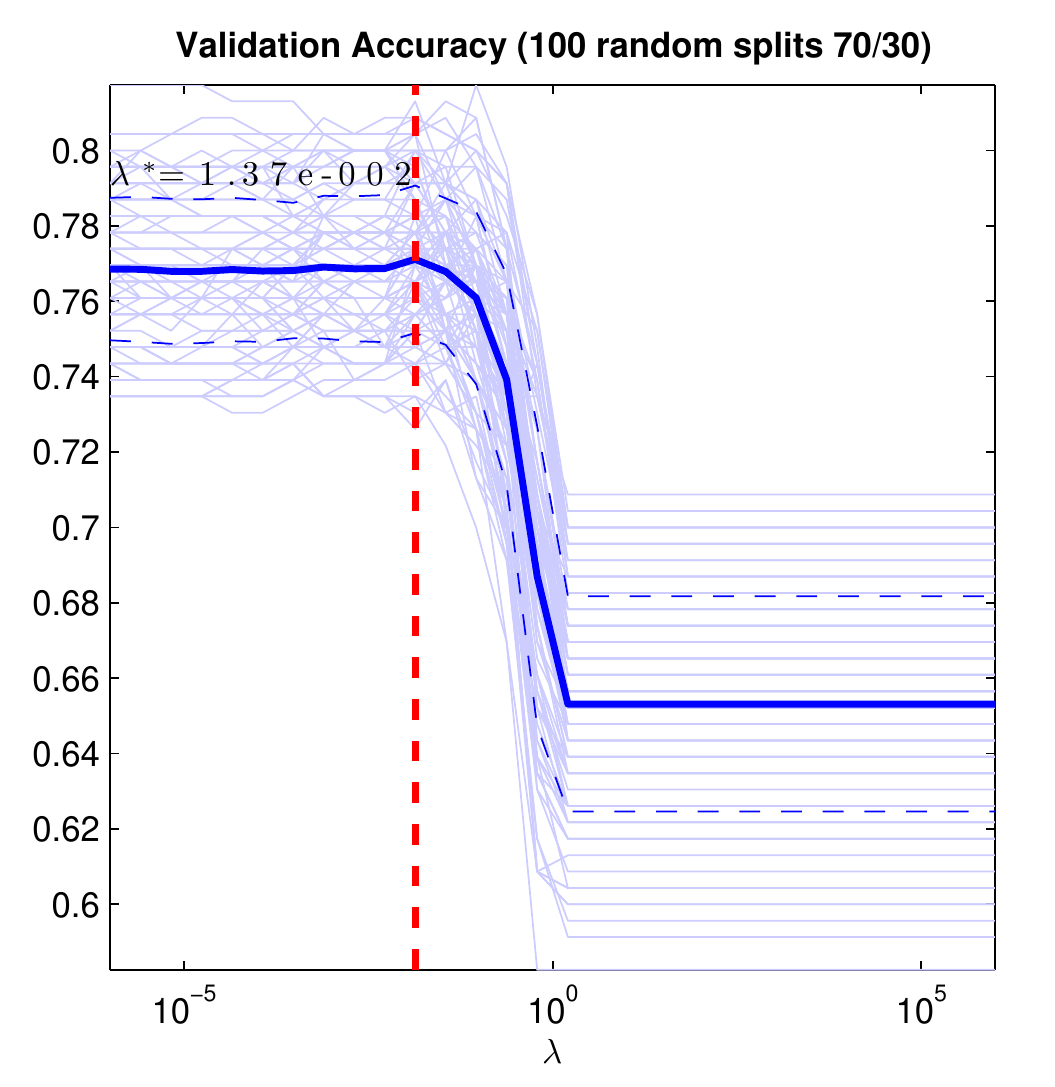}  \includegraphics[width=0.4\linewidth]{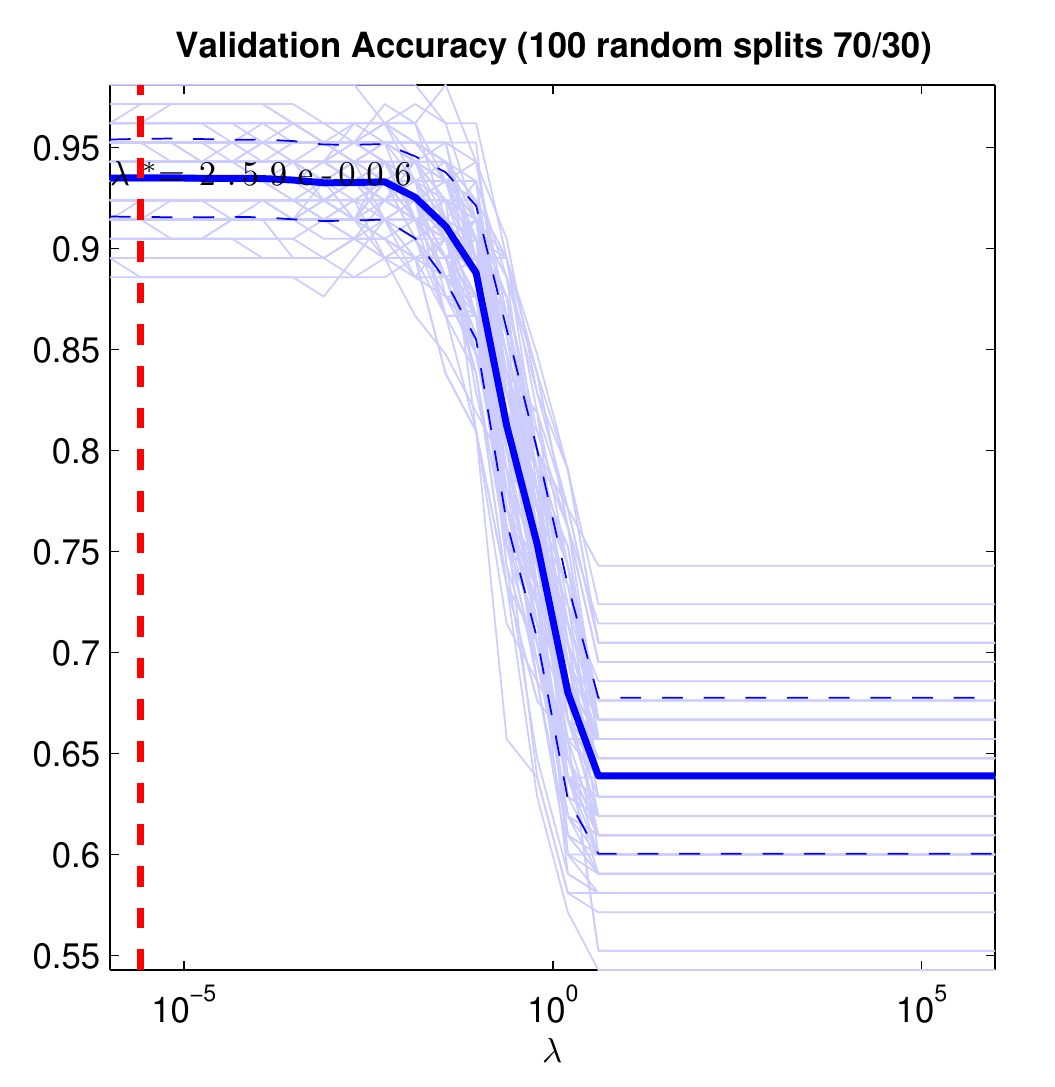}}
 \centerline{\includegraphics[width=0.4\linewidth]{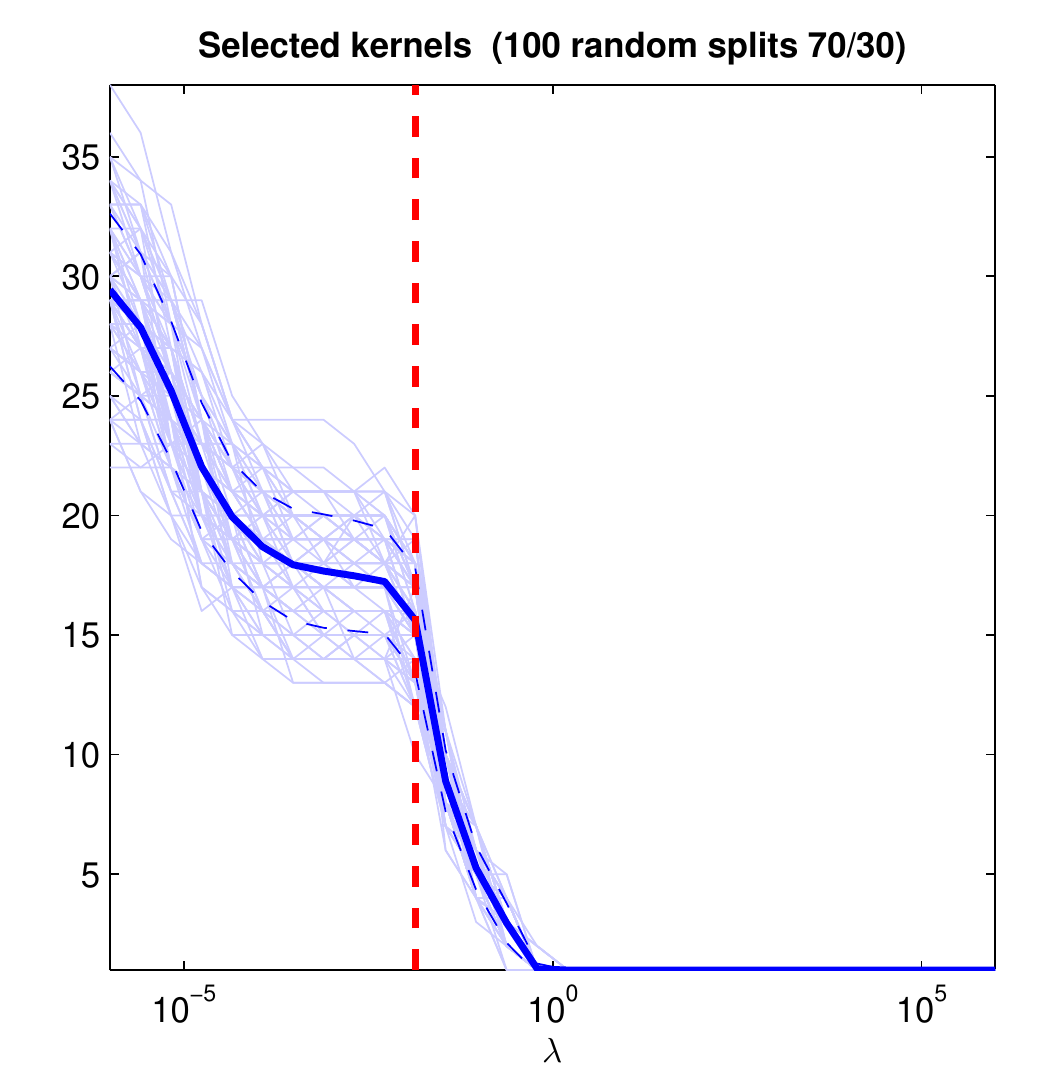}  \includegraphics[width=0.4\linewidth]{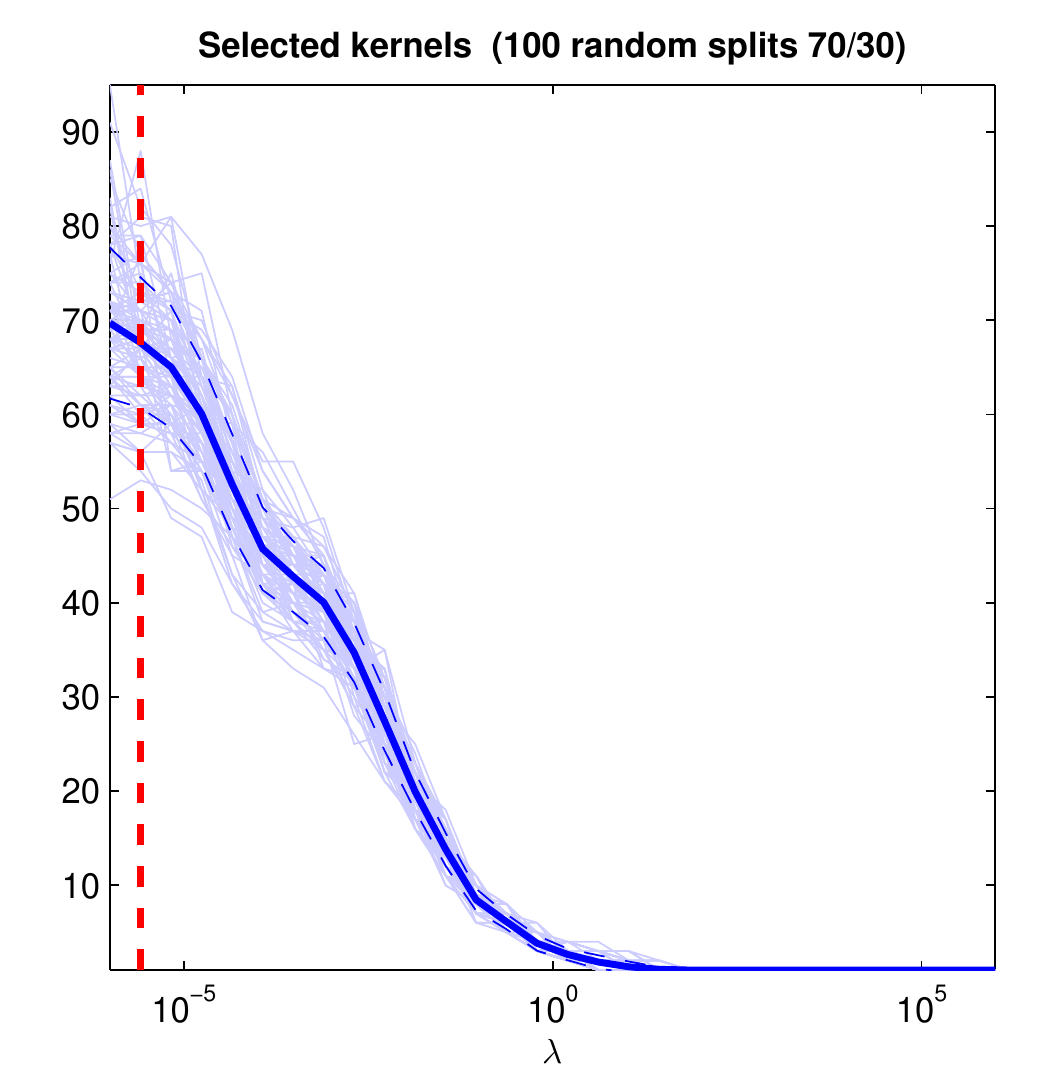}}
 \centerline{\includegraphics[width=0.4\linewidth]{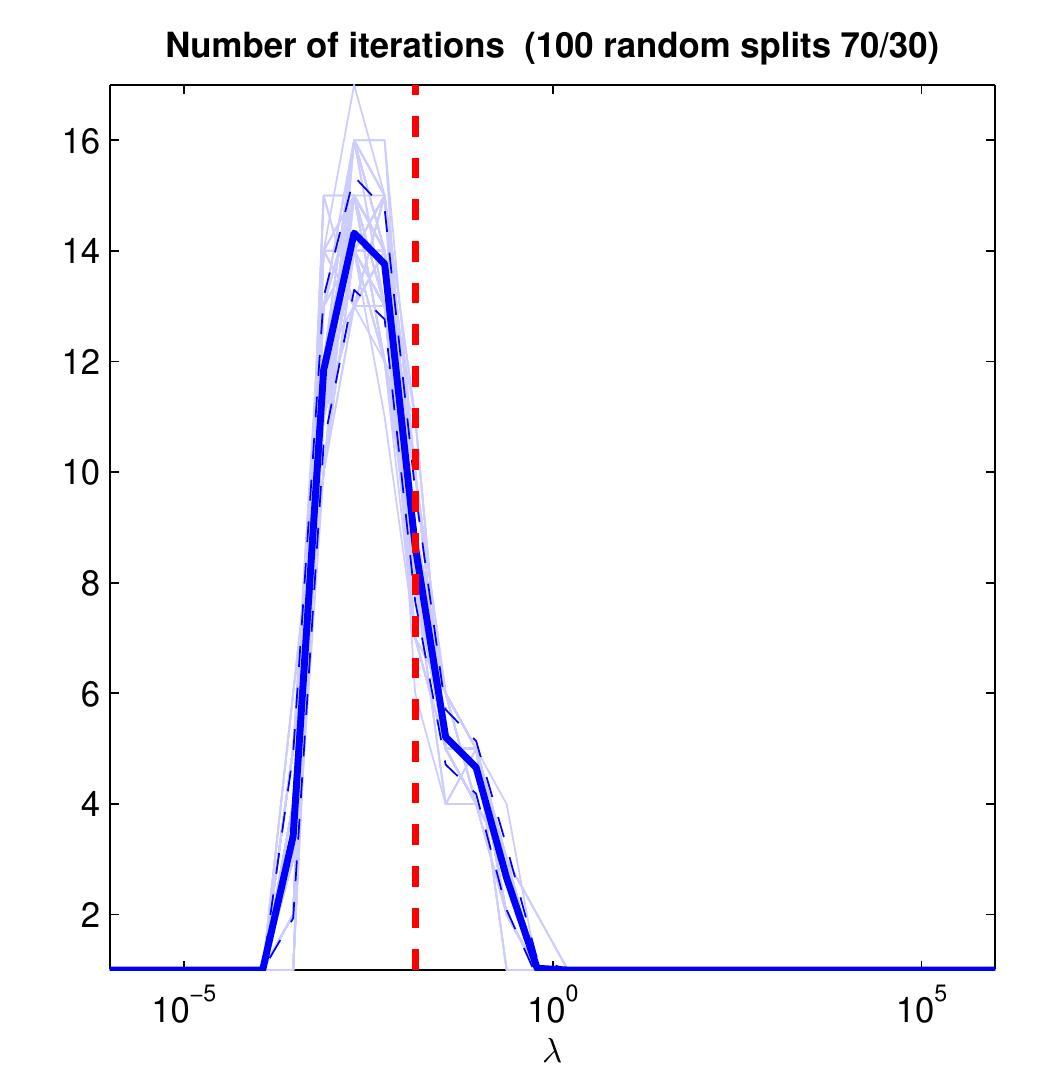}  \includegraphics[width=0.4\linewidth]{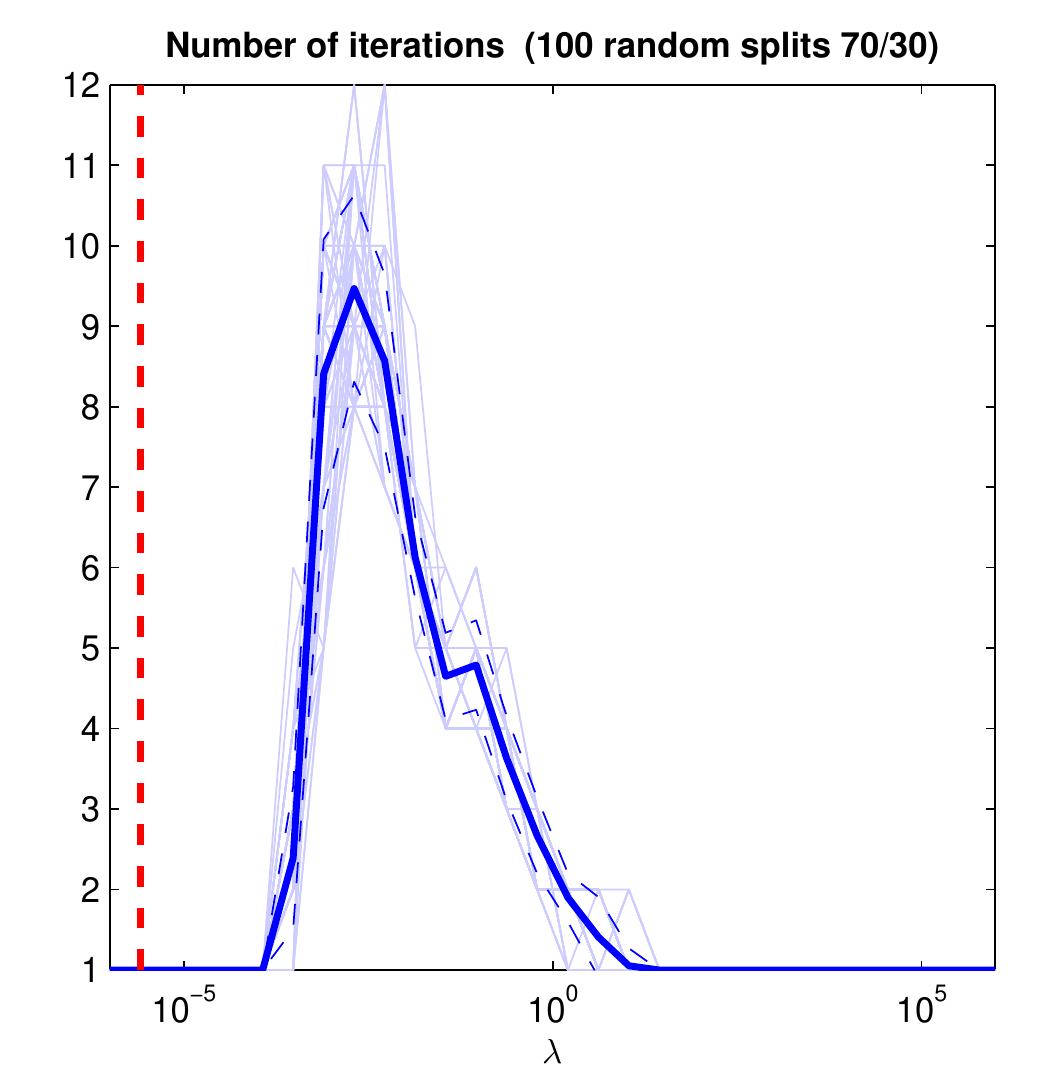}}
  \caption{RLS2 on the \textsc{Pima} (left) and \textsc{Ionosphere} (right) dataset: classification accuracy on the test data (top), number of selected kernels (center), and number of iterations (bottom) along a regularization path.}\label{FIG04}
\end{figure}

\begin{figure}
 \centerline{\includegraphics[width=0.4\linewidth]{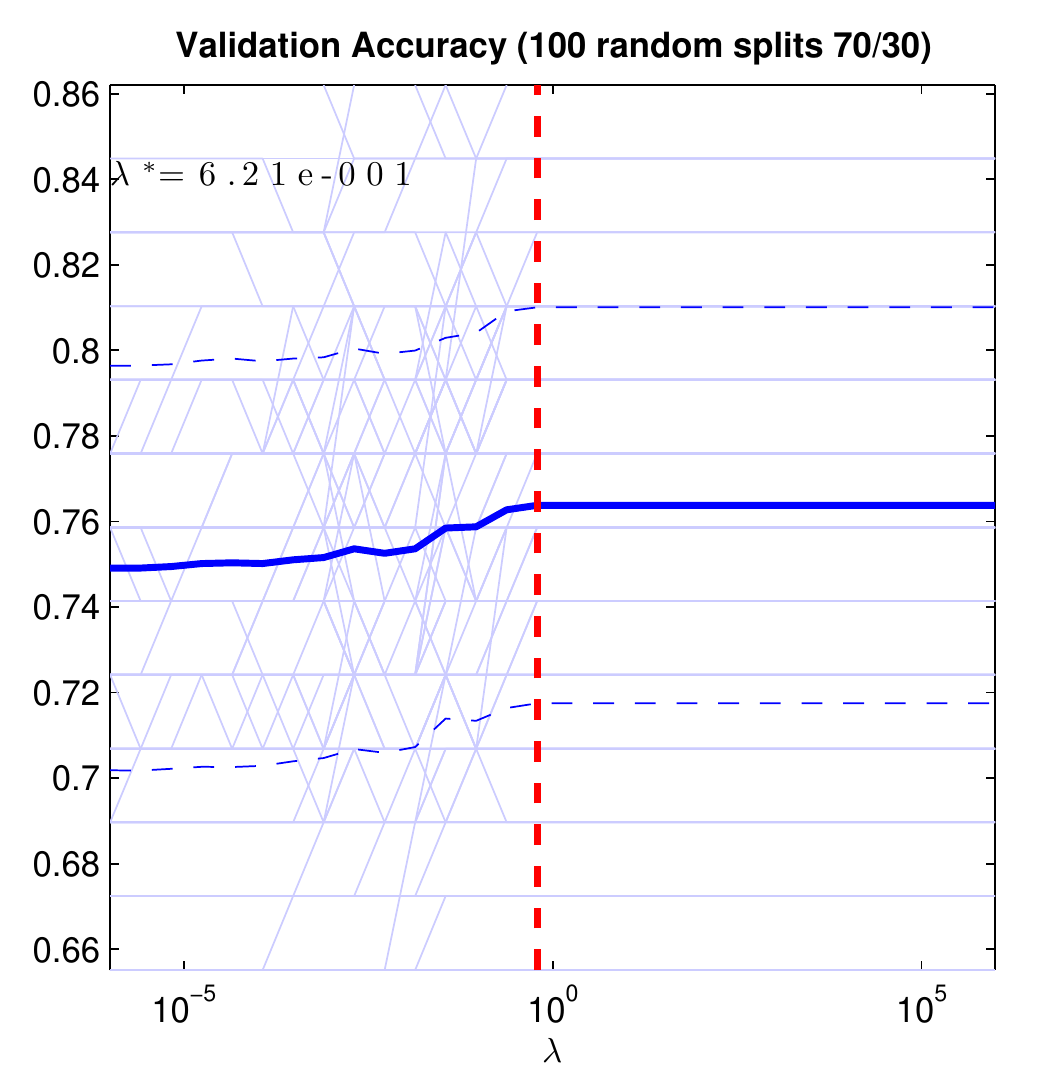}  \includegraphics[width=0.4\linewidth]{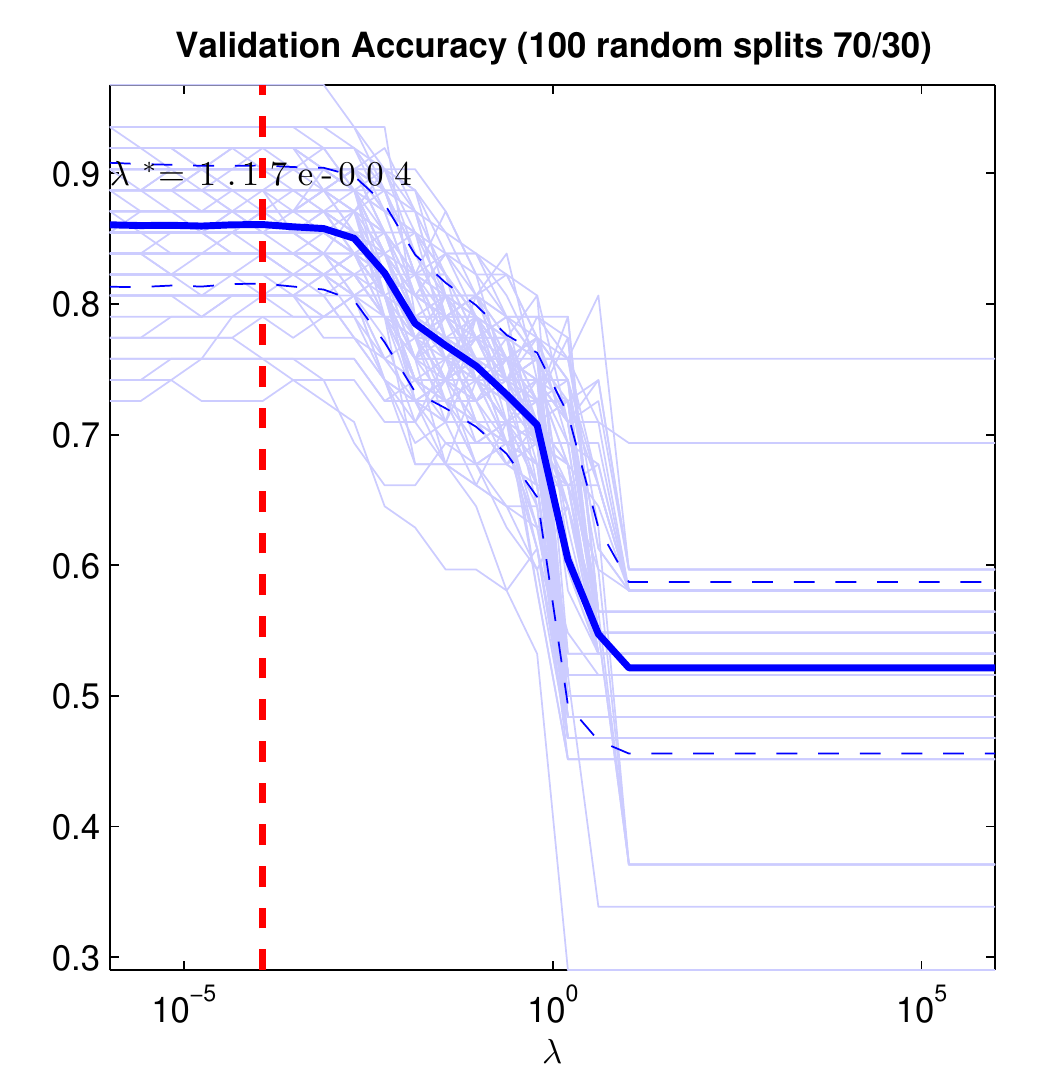}}
 \centerline{\includegraphics[width=0.4\linewidth]{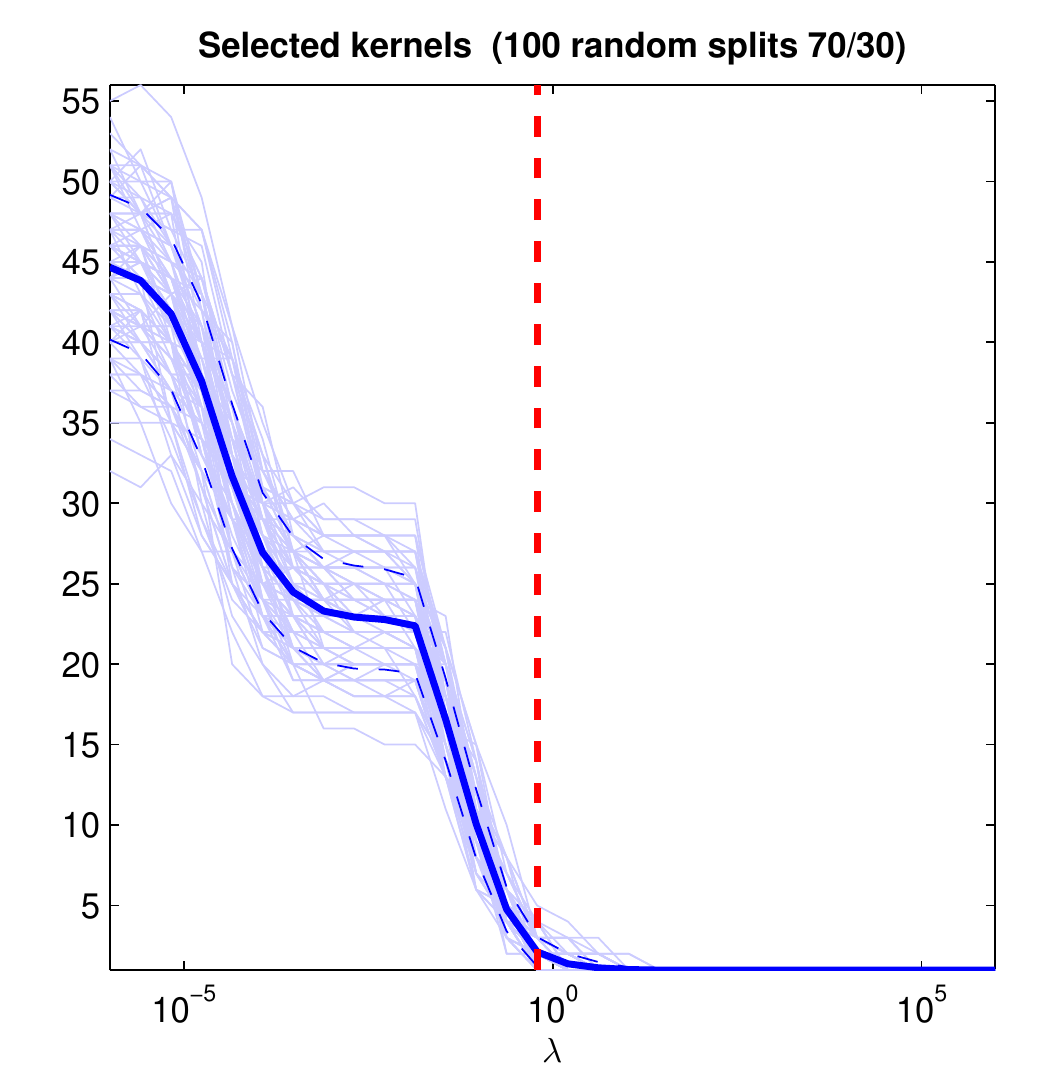}  \includegraphics[width=0.4\linewidth]{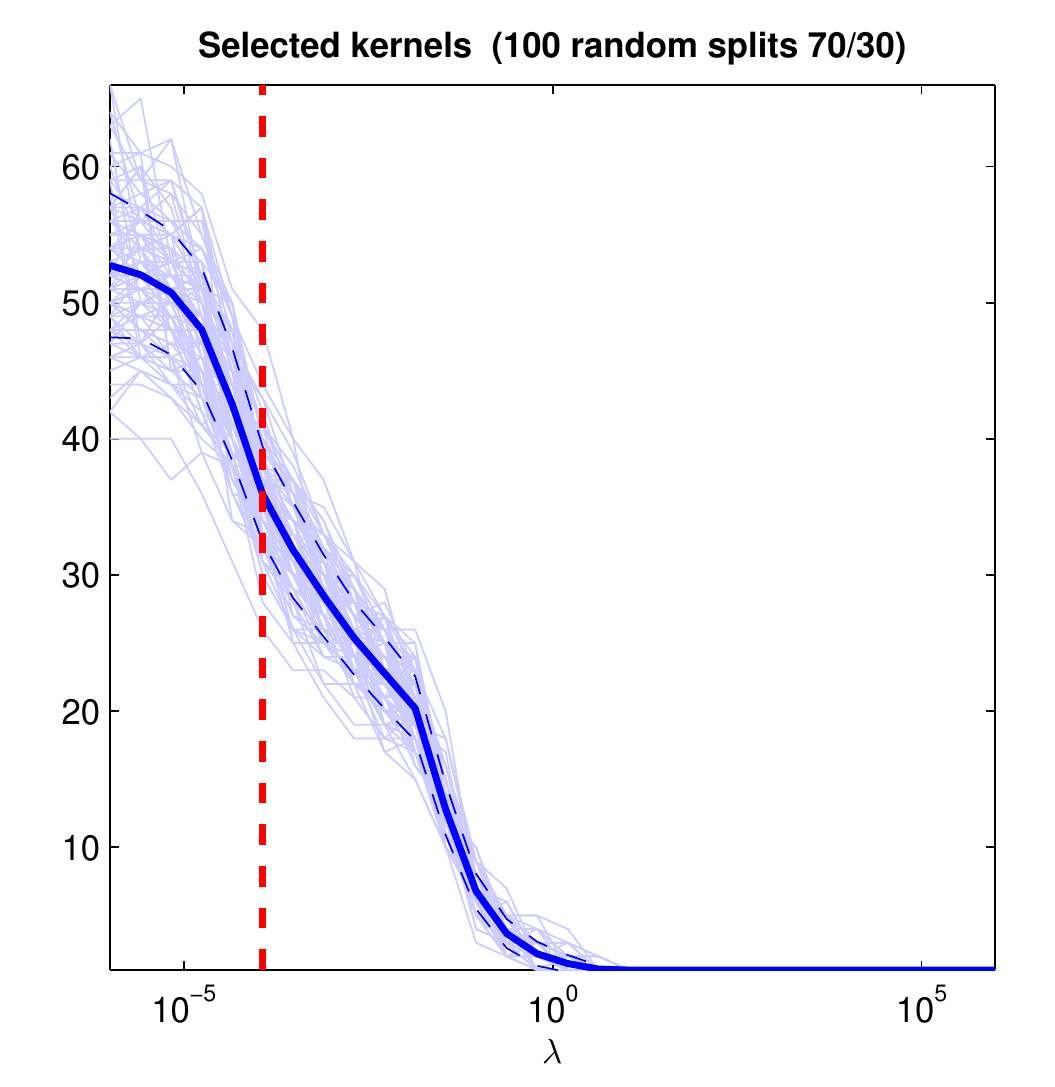}}
 \centerline{\includegraphics[width=0.4\linewidth]{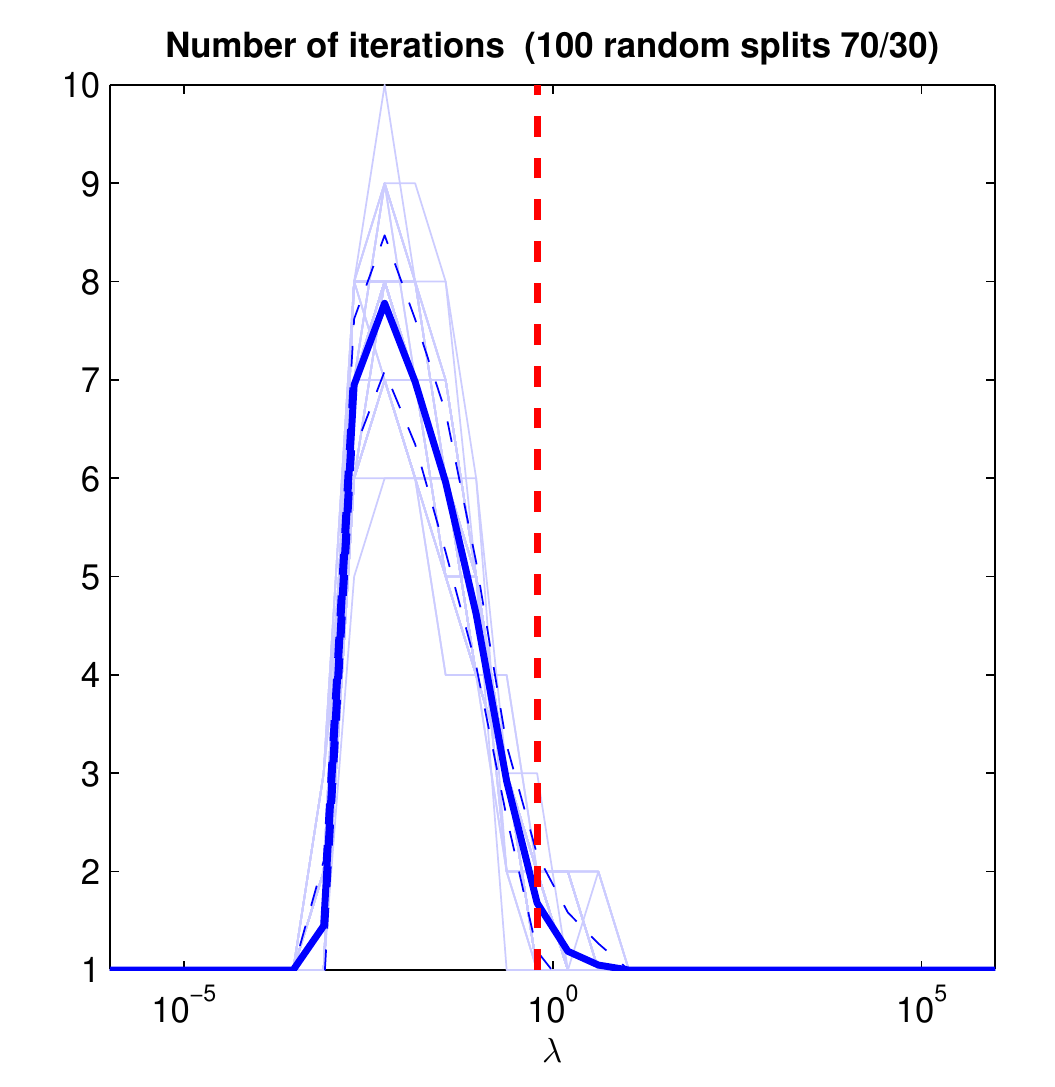}  \includegraphics[width=0.4\linewidth]{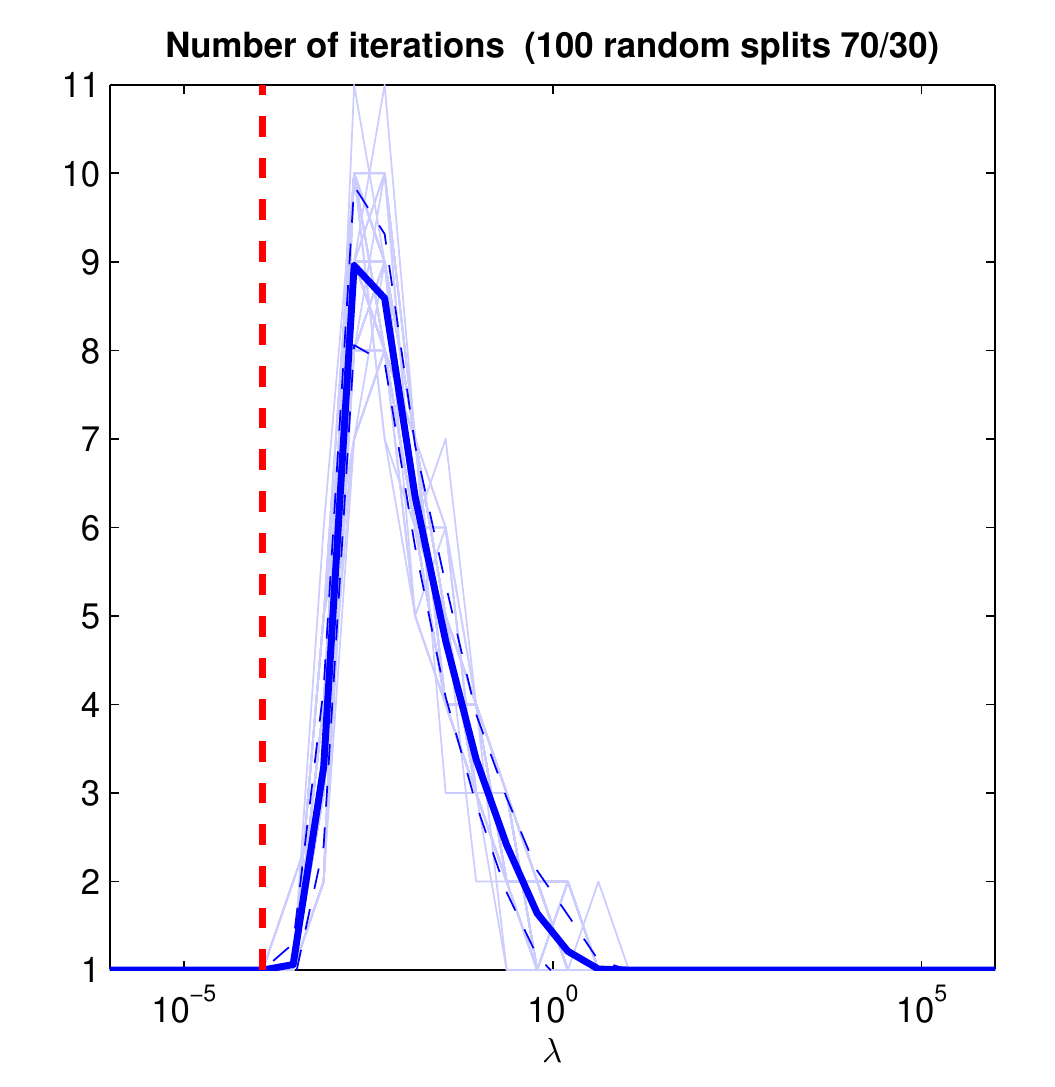}}
  \caption{RLS2 on the \textsc{Wpbc} (left) and \textsc{Sonar} (right) dataset: classification accuracy on the test data (top), number of selected kernels (center), and number of iterations (bottom) along a regularization path.}\label{FIG05}
\end{figure}

\begin{table}
  \caption{RLS2 regression and classification: average and standard deviation of test performance (RMSE for regression, accuracy for classification) over 100 dataset splits. Results with two different training/test ratio are reported: 60/40 (first two columns), and 70/30 (last two columns).}
  \centering
  \begin{tabular}{|l|rr|}
   \hline
   Dataset             & 60/40             & 70/30        \\
   \hline
   \textsc{Auto-mpg}   & $2.79(0.209)$          & $2.72 (0.224)$    \\
   \textsc{Cpu}        & $21.8 (11.3)$          & $21.2(11.9)$      \\
   \textsc{Servo}      & $ 0.755 (0.116)$       & $0.696 (0.152)$   \\
   \textsc{Housing}    & $3.61 (0.465)$         & $3.49 (0.558)$    \\
   \hline
   \textsc{Heart}      & $83.8 (2.98)$          & 84 (3.28)         \\
   \textsc{Liver}      & $69 (3.57)$            & $69.8 (3.79)$     \\
   \textsc{Pima}       & $76.7 (1.92)$          & $77.1 (1.96)$     \\
   \textsc{Ionosphere} & $93.3 (1.83)$          & $93.5 (1.93)$     \\
   \textsc{Wpbc}       & $76.7 (3.71)$          & $76.4 (4.63)$     \\
   \textsc{Sonar}      & $83.6 (3.69)$          & $86.1 (4.52)$     \\
   \hline
  \end{tabular}
  \label{TAB3}
\end{table}

\begin{table}
  \caption{RLS2 regression and classification: number of selected kernels in correspondence with the optimal value of $\lambda$, number of iterations and training time in seconds to compute a regularization path. }
  \centering
  \begin{tabular}{|lrccc|}
  \hline
  Dataset & Split ratio & Kernels & Iterations & Time (s) \\
  \hline
    \textsc{Auto-mpg}
    & 60/40 &  $21.3 (2.64)$ & $78.8 (2.16)$ & $7.2 (1.22)$ \\
    & 70/30 &   $22.1 (2.3)$ & $80.6 (2.17)$ & $17.4 (1.82)$ \\
    \textsc{Cpu}
    & 60/40 & $30.2 (3.65)$ & $46.6 (2.42)$ & $26.4 (13)$ \\
    & 70/30 & $31.9 (3.43)$ & $47 (1.97)$ & $29 (13.8)$ \\
    \textsc{Servo}
    & 60/40 &  $11 (1.37)$ & $71 (2.26)$ & $1.95 (0.214)$ \\
    & 70/30 &  $11.4 (1.8)$ & $73.6 (2.09)$ & $2.37 (0.162)$ \\
    \textsc{Housing}
    & 60/40 & $37.4 (3.95)$ & $90.6 (2.73)$ & $38.7 (1.81)$ \\
    & 70/30 & $38.7 (3.31)$ & $92.4 (2.43)$ & $51.9 (2.07)$ \\
    \hline
    \hline
    \textsc{Heart}
    & 60/40 & $13.7 (2.03)$ & $75.9 (2.33)$ & $4.22 (0.234)$ \\
    & 70/30 & $14.3 (2.12)$ & $77.8 (2.16)$ & $5.83 (0.411)$ \\
    \textsc{Liver}
    & 60/40 & $23.5 (2.87)$ & $54.6 (2.53)$ & $3.56 (0.545)$ \\
    & 70/30 & $15.1 (2.11)$ & $55.1 (2.43)$ & $4.51 (0.399)$  \\
    \textsc{Pima}
    & 60/40 & $14.6 (2.17)$ & $84.7 (2.67)$ & $53.6 (2.23)$ \\
    & 70/30 & $15.6 (2.18)$ & $86.6 (2.33)$ & $70.8 (4.14)$ \\
    \textsc{Ionosphere}
    & 60/40 & $61.6 (6.03)$ & $70.7 (3.13)$ & $16.7 (0.96)$ \\
    & 70/30 & $67.6 (6.92)$ & $73.1 (2.92)$ & $24.2 (2.45)$  \\
    \textsc{Wpbc}
    & 60/40 & $2.29 (0.832)$ & $60.5 (3.03)$ & $4.89 (0.305)$ \\
    & 70/30 & $2.13 (0.906)$ & $60.6 (2.42)$ & $6.85 (0.198)$ \\
    \textsc{Sonar}
    & 60/40 & $45.1 (4.45)$ & $60.5 (2.18)$ & $9.42 (0.191)$ \\
    & 70/30 & $35.9 (3.5)$ & $61.6 (1.9)$ & $11.4 (0.227)$ \\
    \hline
  \end{tabular}
  \label{TAB4}
\end{table}

In this subsection, benchmark experiments on four regression and six classification problems from UCI repository are illustrated (Table \ref{TAB1}). RLS2 has been run on 100 random dataset splits with two different training/test ratios: $60/40$ and $70/30$. For each dataset split, an approximate regularization path with 30 values of $\lambda$ on a logarithmic scale in the interval $\left[10^{-6}, 10^6\right]$ has been computed. To speed-up the regularization path computation, a warm-start technique is employed: the value of $\lambda$ is iteratively decreased, while kernel-expansion coefficients $d_i$ are initialized to their optimal values obtained with the previous value of $\lambda$. Performances are measured by accuracy for classification and RMSE (root mean squared error) for regression. For each dataset split and value of the regularization parameter, the following quantities are computed: prediction performance on the test set, number of selected kernels (number of non-zero $d_i$), training time in seconds and number of iterations to compute the whole regularization path. Datasets have been pre-processed by removing examples with missing features and converting categorical features to binary indicators. For some of the datasets (see Table \ref{TAB1}) input features have been standardized to have zero mean and unitary standard deviation. For classification, output labels are $\pm 1$ and predictions are given by the sign of $(f_2 \circ f_1)$. For regression, an intercept term equal to the mean of training outputs is subtracted to the training data. Basis kernel matrices are pre-computed and the scaling matrix $S$ is chosen according to the rule (\ref{E11}) with transductive scaling.

To better compare the results to similar benchmarks for multiple kernel learning, see e.g. \cite{Rakotomamonjy08}, the same set of basis kernels for all the datasets has been chosen. We remark that such agnostic approach is not representative of a realistic application of the algorithm, in which the choice of basis kernels $\widetilde{K}_k$ should reflect a-priori knowledge about the learning task to be solved. The set of basis kernels contains the following:
\begin{itemize}
  \item Polynomial kernels
  \[
  \widetilde{K}_k(x_1,x_2) = (1+x_1^Tx_2)^{n}
  \]
  \noindent with $n=1,2,3$.
  \item Gaussian RBF kernels
  \[
  \widetilde{K}_k(x_1,x_2) = \textrm{exp}\left(-\gamma \|x_1-x_2\|^2\right),
  \]
  \noindent with 10 different values of $\gamma$ chosen on a logarithmic scale between $10^{-3}$ and $10^{3}$.
\end{itemize}

\noindent Kernels on each single feature and on all the features are considered, so that the number of basis kernels is an affine function of the number of features (recall that categorical features have been converted to binary indicators). More precisely, we have $m = 13 (N+1)$.

All the profiles of test prediction performance, number of kernels and number of iterations for the 70/30 dataset split in correspondence with different values of the regularization parameter are reported in Figures \ref{FIG01}-\ref{FIG05}. From the top plots, it can be seen that test performances are relatively stable to variations of the regularization parameter around the optimal value $\lambda^*$, indicating that RLS2 is robust with respect to the use of different model selection procedures. For regression datasets such as \textsc{Cpu}, \textsc{Servo}, or \textsc{Housing}, optimal performances seems to be reached in correspondence with the un-regularized solution $\lambda \rightarrow 0^+$. Lines in light color are associated with single dataset splits, while thick lines are the averages over different dataset splits. The vertical dotted line corresponds to the value of the regularization parameter with best average test performance. The average number of selected kernels vary quite smoothly with respect to the regularization parameter. For large values of $\lambda$, RLS2  chooses only one basis kernel. For small values of $\lambda$, the number of selected kernels grows and exhibits an higher variability. The bottom plots in Figures \ref{FIG01}-\ref{FIG05} give an idea of the computation burden required by alternate optimization for RLS2 in correspondence with different values of $\lambda$. In correspondence with high values of the regularization parameter, the algorithm converges in a single iteration. This occurs also for the very first value on the regularization path, meaning that the initialization rule is effective. With low values of $\lambda$, RLS2 also converges in a single iteration, since the second layer doesn't change much from an iteration to the next.

Test performances for regression and classification are summarized in Table \ref{TAB3}, where the average and standard deviation with respect to the 100 dataset splits of either RMSE (regression) or accuracy (classification) in correspondence with to the best value of $\lambda$ are reported. Performances of other kernel learning algorithms on some of these datasets can be found in \cite{Lanckriet04, Ong05, Rakotomamonjy08} and references therein. Another benchmark study that might be useful for comparison is \cite{Meyer03}. Comparisons should be handled with care due to the use of different experimental procedures and optimization problems. For instance, \cite{Lanckriet04} uses an 80/20 dataset split ratio, \cite{Ong05} uses 60/40, while \cite{Rakotomamonjy08} uses 70/30. Also, different numbers of dataset splits have been used. Individuating what kind of datasets are better suited to what algorithm is a complex issue, that is certainly worth further investigation. These experiments shows that RLS2 results are competitive and complexity of the model is well controlled by regularization. In particular, state of the art performances are reached on \textsc{Servo}, \textsc{Housing}, \textsc{Hearth}, \textsc{Pima}, \textsc{Ionosphere}.  Finally, it should be observed that, although multiple kernel learning machines have been used as black box methods, the use of basis kernels on single features sometimes also selects a subset of relevant features. Such property is remarkable since standard single-layer kernel methods are not able to perform “embedded” feature selection.

Table \ref{TAB4} reports the average and standard deviation of number of selected kernels in correspondence with the optimal value of $\lambda$, number of iterations and training time needed to compute a regularization path for all the regression and classification datasets studied in this subsection. From the columns of selected kernels, it can be seen that a considerable fraction of the overall number of basis kernels is filtered out by the algorithm in correspondence with the optimal value of the regularization parameter. By looking at the number of iterations needed to compute the path with 30 values of the regularization parameter, one can see that the average number of iterations to compute the solution for a single value of $\lambda$ is in between 1 and 3, indicating that the warm-start procedure is rather effective at exploiting the continuity of solutions with respect to the regularization parameter. As a matter of fact, most of the optimization work is spent in correspondence with a central interval of values of $\lambda$, as shown in the bottom plots of Figures \ref{FIG01}-\ref{FIG05}. Finally, from the last column, reporting average and standard deviation of training times, it can be seen that, with the current implementation of RLS2, regularization paths for all the datasets in this subsection can be computed in less than one minute in the average (see the introduction of this section for experimental details). Although the current implementation of RLS2 has been designed to be efficient, it is believed that there's still considerable margin for further computational improvements. This issue may well be the subject of future developments.

\subsection{RLS2: multi-class classification of microarray data} \label{sec05.3}

\begin{figure}
 \centerline{\includegraphics[width=1\linewidth]{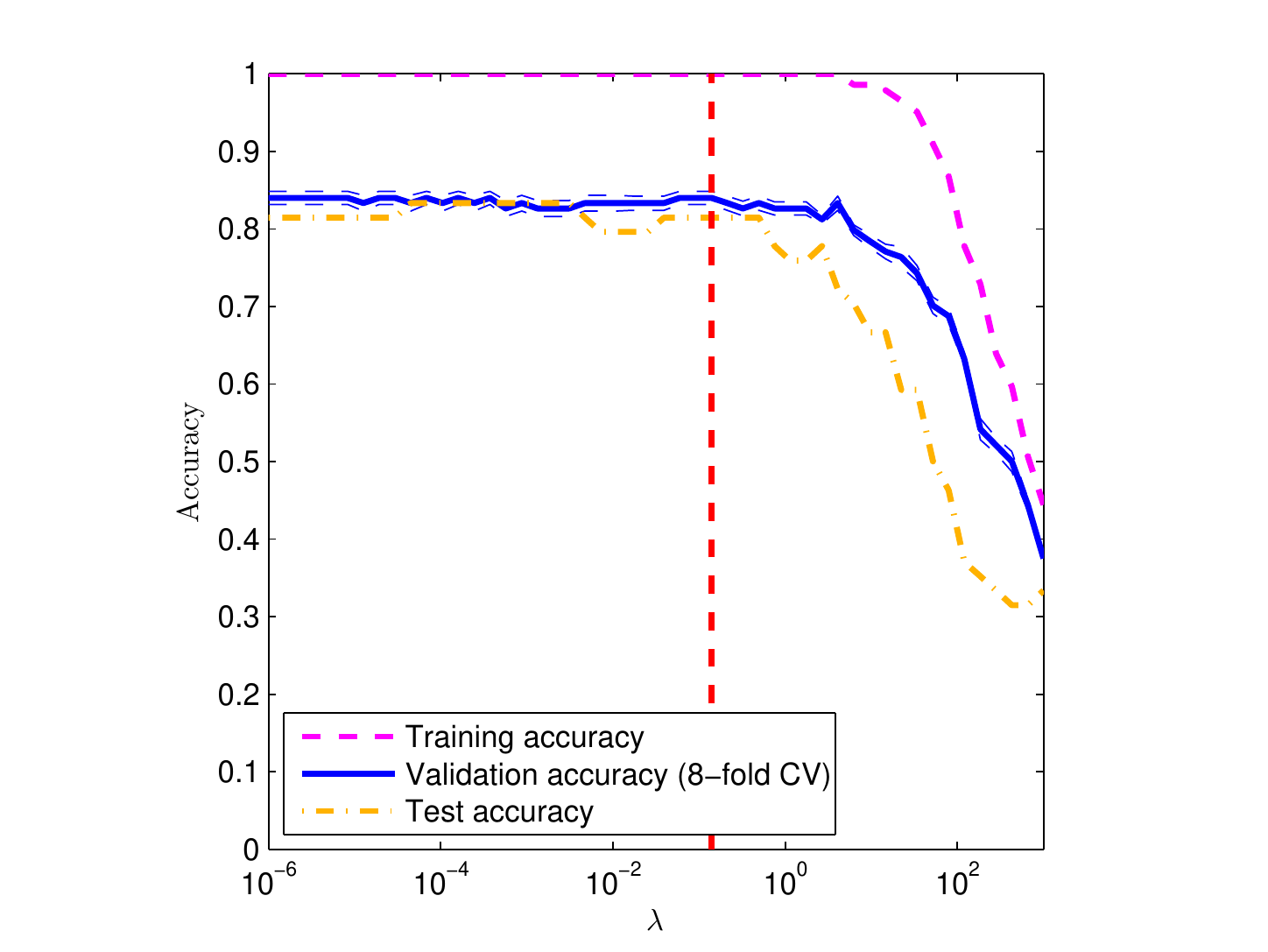} }
 \caption{\textsc{14 Cancers} data: profiles of training error, 8-fold validation error and test error for different values of the regularization parameter. Standard error bands for the validation error are also shown. The vertical line corresponds to the least complex model maximizing the validation accuracy.}\label{FIG12}
\end{figure}

\begin{table}
  \caption{\textsc{14 Cancers} data: average test error (see the text for details) and number of selected genes for different classification algorithms. For RLS2, the number of selected genes is relative to the least complex model maximizing the validation accuracy. Results for methods other than RLS2 are taken from \cite{Hastie08}.}
  \centering
  \begin{tabular}{|rrr|}
   \hline
   Method                                           &Test errors      &   Selected genes    \\
   \hline
   Support Vector Classifier                        &   14.0                    &   16,063                      \\
   L1-penalized multinomial                         &   13.0                    &   269                         \\
   Lasso regression (OVA)                           &   12.5                    &   1,429                       \\
   L2-penalized discriminant analysis               &   12.0                    &   16,063                      \\
   Elastic-net penalized multinomial                &   11.8                    &   384                         \\
   Linear RLS2 (OVA)                                &   \textbf{9.8}            &   855                         \\
   \hline
  \end{tabular}
  \label{TAB5}
\end{table}

RLS2 can be applied to multi-class classification problems by solving several binary classification problems and combining their outcomes. A possible way to combine binary classifiers is the OVA (one versus all) approach, in which each class is compared to all the others and test labels are assigned to the class maximizing the confidence (the real-valued output) of the corresponding binary classifier.

Linear RLS2 with OVA has been applied to the \textsc{14 Cancers} dataset \cite{Ramaswamy01}, a delicate multi-class classification problem whose goal is to discriminate between 14 different types of cancer, on the basis of microarray measurements of 16063 gene expressions. Gene measurements and type of cancer (labels) are available for 198 patients, the dataset being already divided in a training set of 144 patients, and a test set of 54 patients. Another important goal in this problem is to individuate a small subset of genes which is relevant to discriminate between the different kind of cancer. \cite{Hastie08} reports several results for these data using a variety of classification methods. Algorithms such as the Support Vector Classifier uses all the genes to compute the classification boundaries, while others such as Lasso or Elastic Net are also able to select a subset of relevant genes. Since the feature selection experiment in subsection \ref{sec05.1} suggests that RLS2 may be very efficient at selecting relevant features from noisy examples, a microarray dataset seems to be an appropriate choice for testing the algorithm.

Gene expressions for each patient have been firstly standardized to have zero mean and variance one. For each binary classifier, coefficients $s_i$ are chosen as $s_i = \left(\sigma_+^2+\sigma_-^2\right)^{-1/2}$, where $\sigma_+^2$ and $\sigma_-^2$ are the within-class sample variances computed using all the training data. Such scaling gives more weight to genes whose expressions exhibits small within-class variability, and seems to slightly improve classification performances. A validation accuracy profile has been computed using stratified 8-fold cross validation, where the folds are organized to approximately preserve the class proportions\footnote{We thank Trevor Hastie for kindly providing the folds used in their experiments.}. For the final model, we pick the highest value of $\lambda$ maximizing the validation accuracy. Figure \ref{FIG12} reports the profiles of training accuracy, cross-validation accuracy with corresponding standard error bands, and test accuracy for 50 logarithmically spaced values of the regularization parameter. Table \ref{TAB5} reports the number of test errors and selected genes in correspondence with the value of $\lambda$ chosen in the validation phase, for RLS2 and other methods from \cite{Hastie08}. Test errors in Table \ref{TAB5} are averages of test errors for different classifiers associated with all the different values of the regularization parameter that maximizes the cross-validation score (this explains the presence of non-integer values). For linear RLS2, such procedure yields a value of about 9.8. In correspondence with the least complex model maximizing the cross-validation accuracy, one obtain 10 test errors using 855 genes. Although the test set size is too small to draw significative conclusions from this comparison, linear RLS2 seems to work rather well on this problem and achieve the best test performances. Such good performance also confirm  effectiveness of the OVA multi-class approach for RLS2.

\section{Conclusions} \label{sec06}

The connection between learning with a two-layer network and the problem of learning the kernel has been analyzed. While architectures with more than one layer are justified by a representer theorem, an alternative perspective to look at the problem of kernel learning is proposed. Such perspective makes clear that these two methodologies aim both at increasing the approximation power of standard single layer methods by using machines that can adaptively select functions with a variety of shapes when little prior knowledge is available. In particular, the multiple kernel learning framework is shown to be an important specific case of a more general two-layer architecture. We also introduce RLS2, a new method to perform multiple kernel learning based on regularization with the square loss function and alternate optimization. RLS2 exhibits state of the art performances on several learning problems, including multi-class classification of microarray data. An open source set of MATLAB scripts for RLS2 and linear RLS2 is available at \url{http://www.mloss.org} and also includes a Graphic User Interface.

%\acks{}

\newpage

\appendix

\section*{Appendix (proofs)}

\noindent \textbf{Proof of Theorem \ref{THM1}}

For any fixed $f_1$, let $z_i := f_1(x_i)$. By fixing an optimal $f_1$, Problem \ref{PBM01} can be written as a function of $f_2$ alone as
\[
\min_{f_2 \in\mathcal{H}_2}\left(\sum_{i=1}^{\ell}L_i\left(f_2(z_i)\right) + R_2(\|f_2 \|_{\mathcal{H}_2})\right).
\]

\noindent By standard representer theorems for vector valued functions (see \cite{Micchelli05} and the remark on monotonicity in \cite{Scholkopf01} after Theorem 1), there exists an optimal $f_2$ in the form
\[
f_2(z) = \sum_{i=1}^{\ell} K^2(z_i,z)b_i.
\]

\noindent Then, by fixing on optimal $f_2$ in this form, Problem \ref{PBM01} can be written as a function of $f_1$ alone as
\[
\min_{f_1 \in \mathcal{H}_1 } \left(\sum_{i=1}^{\ell}\widetilde{L}_i\left(f_1(x_i)\right) +R_1(\|f_1 \|_{\mathcal{H}_1})\right),
\]
\noindent where $\widetilde{L}_i\left(z\right) := L_i\left(f_2(z)\right)$. Notice that the new loss functions $\widetilde{L}_i$ depends on $f_2$. Again, by the single-layer representer theorem the finite kernel expansion for $f_1$ follows. Finally, it is immediate to see that the overall input-output relation $f_2 \circ f_1$ can be written as in (\ref{E02}).

\noindent \textbf{Proof of Lemma \ref{LEM1}}

By linearity of $f_2$, it is immediate to see that (\ref{E15}) holds. It follows that
\[
\sum_{i=1}^{\ell}L_i\left((f_2\circ f_1)(x_i)\right) = \sum_{i=1}^{\ell}L_i\left((f_2^*\circ f_1^*)(x_i)\right).
\]
\noindent In addition,
\[
\frac{\alpha}{2} \| f_1^* \|^2_{\mathcal{H}_1} = \frac{\alpha}{\beta^2 2} \| f_1 \|^2_{\mathcal{H}_1}, \qquad \gamma \cdot I\left(\frac{\| f_2^* \|_{\mathcal{H}_2}}{\beta} \right) = I\left(\| f_2 \|_{\mathcal{H}_2} \right).
\]
\noindent In the last equation, we exploit the fact that $\gamma I(x) = I(x)$, for any positive $\gamma$, a property that is satisfied only by indicator functions. The thesis follows by letting $\lambda = \alpha/\beta^2$.

\noindent \textbf{Proof of Theorem \ref{THM2}}

Problem \ref{PBM02} is a specific instance of Problem \ref{PBM01}. The functional to minimize is bounded below, lower semi-continuous and radially-unbounded with respect to $(f_1,f_2)$. Existence of minimizers follows by weak-compactness of the unit ball in $\mathcal{H}_1$ and $\mathcal{H}_2$. By Theorem \ref{THM1}, there exists an optimal $f_1$ in the form
\[
f_1(x) = \sum_{j=1}^{\ell} K^1(x,x_j)a_{j} = \sum_{j=1}^{\ell} \textrm{diag}\left\{\widetilde{K}_{1}(x,x_j), \ldots, \widetilde{K}_{m}(x,x_j) \right\}a_{j}.
\]
\noindent Introduce the matrix $A \in \mathbb{R}^{m \times \ell}$ whose rows are denoted by $(c^i)^T$ and whose columns are $a_j$. Then, the $i$-th component of $f_1$ can be written as:
\[
f_1^i(x) = \sum_{j=1}^{\ell} c^i_{j} \widetilde{K}_{i}(x,x_j).
\]

\noindent By Theorem \ref{THM1}, there exists an optimal $f_2$ such that
\[
f_2(z) = \sum_{j=1}^{\ell} b_j K^2(f_1(x_j),z) = \sum_{j=1}^{\ell}b_j z^T S f_1(x_j) = z^TS \sum_{j=1}^{\ell}b_{j}f_1(x_j) = z^TS w,
\]
\noindent where
\[
w :=  \sum_{j=1}^{\ell}b_{j}f_1(x_j).
\]

\noindent Letting matrices $R^k \in \mathbb{R}^{\ell \times \ell}$ like in (\ref{E06}) and $Q$ as in (\ref{E10}), Problem \ref{PBM02} can be rewritten as
\[
\min_{c^1, \ldots c^{m} \in \mathbb{R}^{\ell}, w \in \mathbb{R}^{m}}\left[Q\left(\sum_{k=1}^{m} w_k R^k c^k\right) + \frac{\lambda}{2}\sum_{k=1}^{m} \frac{(c^{kT} R^k  c^k)}{s_k}\right], \quad \textrm{ subject to } \quad w^T S w \leq 1.
\]

\noindent By optimizing with respect to vectors $c^i$, we have
\[
0 \in - s_iw_i R^i  \partial Q\left(\sum_{k=1}^{m}w_k R^k c^k\right)+ \lambda R^i c^i,
\]

\noindent where $\partial$ is the sub-differential of a convex function \cite{Rockafellar70}. Now, letting
\[
c  \in \frac{1}{\lambda}\partial Q\left(\sum_{k=1}^{m}w_k R^k c^k\right),
\]
\noindent we obtain
\[
c^i = s_i w_i c.
\]

\noindent Letting $d_i := s_i w_i^2$ and $R := \sum_{i=1}^{m} d_i R^i$, Problem \ref{PBM02} boils down to Problem \ref{PBM07}. By Theorem \ref{THM1} again, the overall input-output relation can be written as in equation (\ref{E02}), where the kernel $K$ satisfy
\begin{eqnarray*}
K(x_1,x_2) & = & K^2(f_1(x_1),f_1(x_2)) = f_1(x_1)^T S f_1(x_2) = \sum_{i=1}^{m}s_i f_1^i(x_1)f_1^i(x_2)\\
 & = & \sum_{i=1}^{m}s_i w_i^2 \sum_{j_1=1}^{\ell}\sum_{j_2=1}^{\ell}c_{j_1}c_{j_2} \widetilde{K}_i(x_{j_1},x_1)\widetilde{K}_i(x_{j_2},x_2)\\
 & = & \sum_{i=1}^{m} d_i K_i(x_1,x_2),
\end{eqnarray*}

\noindent and $K_i$ are as in (\ref{E03}).

\noindent \textbf{Proof of Lemma \ref{LEM2}}
Problem \ref{PBM03} can be rewritten as
\begin{equation}\label{E08}
\min_{z \in \mathbb{R}^{\ell}, R \in \mathbb{S}^{+}_{m}}\left(Q(z) + \frac{\lambda}{2}z^T R^{\dag}z \right),
\end{equation}
\noindent subject to (\ref{E06}), where $\mathbb{S}^{+}_{m}$ denotes the cone of $m \times m$ positive semi-definite matrices. This problem can be seen to be jointly convex in $(z,R)$ using an argument due to \cite{Kim08}: the term $z^T R^{\dag}z$ is a \emph{matrix-fractional function} (see e.g. \cite{Boyd04}, Example 3.4), which is a jointly convex function of the pair $(z,R)$. This easily follows by noticing that its epi-graph is a convex set:
\[
z^T R^{\dag}z \leq \alpha \quad \Leftrightarrow \quad \left(
                                                                 \begin{array}{cc}
                                                                   \alpha & z^T \\
                                                                   z & R \\
                                                                 \end{array}
                                                               \right) \in \mathbb{S}^{+}_{m+1}.
\]
\noindent Since $Q$ is a convex function, the overall functional (\ref{E08}) is convex. Minimization in (\ref{E08}) subject to linear constraints (\ref{E06}) is thus a convex optimization problem. Since $R$ is a linear function of $d$, Problem \ref{PBM03} is also convex.

To prove (\ref{E07}), assume that $(z^*,d^*)$ is an optimal pair for Problem \ref{PBM03}. Without loss of generality, we can assume $d^* \neq 0$. Indeed, if there's an optimal solution with $d^* = 0$, then $z = 0, d \neq 0$ is optimal as well. Now, let $\gamma := \sum_{i=1}^{m}d_i^* $, and notice that $0 < \gamma \leq 1$. Introducing the new pair $\left(z,  d\right) = \left(z^*, d^*/\gamma\right)$, the value of the objective functional in correspondence with $\left(z,  d\right)$ is
\[
Q(z) + \frac{\lambda}{2}z^T R^{\dag}(d) z= Q(z^*) + \frac{\lambda\gamma}{2}(z^{*})^T R^{\dag}(d^*) z^* \leq Q(z^*) + \frac{\lambda}{2}(z^*)^T R^{\dag}(d^*) z^*,
\]
\noindent so that the new pair is optimal as well.

\noindent \textbf{Proof of Lemma \ref{LEM3}}

By Lemma \ref{LEM2}, minimization with respect to $d$ can be restricted to the standard simplex $\Delta_m$. For any fixed $d$, the functional of Problem \ref{PBM04} is a convex quadratic function of $c$. If $c^*(d)$ satisfy equation (\ref{E09}), then the gradient of the objective functional with respect to $c$ is zero in $c^*$, meaning that $c^*$ is optimal. Dropping the dependence on $d$, equation (\ref{E09}) can be rewritten as
\[
y-Rc^* =  \lambda c^*.
\]
\noindent In correspondence with such optimal $c^*$, we have
\[
\frac{1}{2}\left\|y- R c^*\right\|^2 + \frac{\lambda}{2}c^{*T} R c^* =  \frac{\lambda^2}{2}\left\|c^* \right\|^2+\frac{\lambda}{2}c^{*T}\left(y - \lambda c^*\right) = \frac{\lambda}{2}y^Tc^{*}.
\]

\noindent \textbf{Proof of Lemma \ref{LEM4}}

By Lemma \ref{LEM2}, minimization with respect to $d$ can be restricted to the standard simplex $\Delta_m$. In addition, we have
\begin{eqnarray*}
\frac{1}{2}\left\|y- R c\right\|^2 + \frac{\lambda}{2}c^T R c & = & \frac{1}{2}\left\|u- R c+\frac{\lambda c}{2}\right\|^2 + \frac{\lambda}{2}c^T R c\\
& = & \frac{1}{2}\left\|u- R c\right\|^2 + \frac{\lambda}{2}c^T\left(\frac{\lambda}{2}c+u\right) = \frac{1}{2}\left\|u- R c\right\|^2 + \frac{\lambda}{2}c^Ty,
\end{eqnarray*}
\noindent where $\lambda c^Ty /2$ does not depend on $R$ (and thus does not depend on $d$). Now, recalling that
\[
R(d) = \sum_{i=1}^{m} d_iR^i,
\]
\noindent we have
\[
R c = \sum_{i=1}^m  d_i R^i c = \sum_{i=1}^m d_i v_i = V d.
\]

\noindent \textbf{Proof of Lemma \ref{LEM5}}

From equation (\ref{E09}), we have
\[
c_{\infty} = \lim_{\lambda\rightarrow +\infty}\left(R(d) +\lambda I\right)^{-1} y = 0.
\]

\noindent Since $R(d)$ is a continuous function of $d$ defined over the compact set $\Delta_m$, by fixing any matrix norm $\| \cdot \|$ there exists a sufficiently large $\underline{\lambda}$ such that
\[
\max_{d \in \Delta_m}\|R(d)\| <  \underline{\lambda}.
\]

\noindent For $\lambda > \underline{\lambda}$, the expansion
\[
\left(R(d) / \lambda+I\right)^{-1} = I-R(d) /\lambda + o(1/\lambda^2),
\]

\noindent holds. By Lemma \ref{LEM3}, it follows that
\begin{eqnarray*}
d^*(\lambda) & = & \arg\min_{d \in \Delta_m} y^{T} \left(R(d) / \lambda+I\right)^{-1} y\\
& = & \arg\min_{d \in \Delta_m}\left[\lambda \|y\|^2/2-(y^T R(d) y) + o(1/\lambda)\right]\\
& = & \arg\min_{d \in \Delta_m}\left[\|y\|^2/2-(y^T R(d) y)/\lambda + o(1/\lambda^2)\right]\\
& = & \arg\max_{d \in \Delta_m}\left[y^T R(d) y - o(1/\lambda) \right].
\end{eqnarray*}

\noindent Hence, $d_{\infty} = \lim_{\lambda \rightarrow +\infty} d^*(\lambda)$ solves the following linear program
\[
\max_{d \in \Delta_m}\sum_{i=1}^{m}d_i (y^T  R^i y).
\]
\noindent Then, it is easy to see that $d = e_k,  k \in \arg\max_{i=1,\ldots,m} y^T  R^i y$, is an optimal solution of the linear program, where $k$ is any index maximizing the “kernel alignment” $y^T R^i y$.

\noindent \textbf{Proof of Lemma \ref{LEM6}}

When basis kernel are chosen as in (\ref{E12}), $f(x)$ can be written as in (\ref{E16})-(\ref{E17}), and we have
\[
R(d) = H \Gamma H^T.
\]
\noindent By letting $z := H^T c$, it follows that
\[
Rc = H \Gamma z = \widetilde{H} \widetilde{z}
\]
\[
c^TRc = c^T H \Gamma H^T c =  \|\Gamma^{1/2} z\|^2 = \|\widetilde{\Gamma}^{1/2} \widetilde{z}\|^2
\]
\noindent Hence, Problem \ref{PBM04} reduces to Problem \ref{PBM08}.

\end{document}